\documentclass{article}
\usepackage{iclr2027_conference,times}
\usepackage[T1]{fontenc}

\usepackage{amsmath,amsfonts,bm}

\def\eqref#1{equation~\ref{#1}}

\def\1{\bm{1}}

\DeclareMathAlphabet{\mathsfit}{\encodingdefault}{\sfdefault}{m}{sl}
\SetMathAlphabet{\mathsfit}{bold}{\encodingdefault}{\sfdefault}{bx}{n}

\usepackage{graphicx}
\usepackage{booktabs}
\usepackage{multirow}
\usepackage{wrapfig}
\usepackage{placeins}
\usepackage{xcolor}
\usepackage{soul}
\sethlcolor{yellow!45}
\usepackage{colortbl}
\usepackage{hyperref}
\usepackage{url}
\hypersetup{hidelinks}

\title{How Learning Governs Unlearning across the Memorization--Generalization Spectrum}
\author{%
Hwiyeong Lee\textsuperscript{1}%
\thanks{Equal contribution\quad
  \textsuperscript{\ensuremath{\dagger}}%
  Work done while at Hanyang University}%
\kern0.4em\footnotemark[2]\quad
Hyelim Lim\textsuperscript{2}\footnotemark[1]\quad
Ingyu Bang\textsuperscript{2}\quad
Hoki Kim\textsuperscript{3}%
\setcounter{footnote}{2}%
\thanks{Corresponding authors}\quad
Taeuk Kim\textsuperscript{2}\footnotemark[3]
\\[0.5em]
\textsuperscript{1}Seoul National University\quad
\textsuperscript{2}Hanyang University\quad
\textsuperscript{3}Chung-Ang University
\\[0.3em]
\texttt{hyglee@snu.ac.kr}\quad
\texttt{hokikim@cau.ac.kr} \\
\texttt{\{yomilimi,ingyu1008,kimtaeuk\}@hanyang.ac.kr}
}
\iclrfinalcopy

\begin{document}
\maketitle
\lhead{Preprint}
\begin{abstract}

While unlearning seeks to negate undesired capabilities acquired through learning, 
little research has examined how the way models learn shapes their subsequent unlearning.
In this paper, we investigate this connection from the perspectives of memorization and generalization, the two most representative yet competing strategies that models employ during training.
We first classify memorization- and generalization-heavy models using grokking in modular addition and compare their responses to unlearning, showing that the latter suffer greater retain damage, i.e., a larger performance drop on the retain set.
Furthermore, we conduct a finer-grained analysis by introducing bucketed modular addition, in which the respective contributions of the two strategies can be explicitly controlled across the memorization--generalization spectrum. 
In this setup, we reaffirm that the same trend persists and is nearly monotonic.
We further demonstrate that this relationship also holds in LLM unlearning across verbatim and factual recall settings.
Finally, we provide two practical insights for developing better unlearning methods, highlighting the importance of accounting for learning dynamics in unlearning. 
\end{abstract}

\section{Introduction}
\label{sec:introduction}

Trained on vast corpora, pre-trained networks such as large language models (LLMs) can contain private information or copyrighted material that data owners may later request be removed \citep{carlini2021extracting,eldan2024whos}. 
Machine unlearning aims to eliminate the knowledge a model has acquired from such data while preserving its general capabilities \citep{bourtoule2021unlearning}. 
In practice, however, existing unlearning methods rarely achieve this balance, typically incurring the loss of unrelated knowledge and skills \citep{shi2025muse}, an effect we refer to as \textbf{retain damage}.

In this work, we investigate how learning dynamics shape the difficulty of unlearning and, in turn, the amount of retain damage it entails.
Our intuition is that, since unlearning attempts to reverse training, the strategies a model employs to fit its data should determine how readily it later yields to unlearning. 
Neural networks are known to rely on the two most representative learning mechanisms, \textbf{memorization} and \textbf{generalization}, to different extents \citep{bengio2013representation,liu2022towards}.
We thus regard the two as endpoints of a memorization--generalization spectrum, locate a trained model on it by its relative reliance on each, and measure how retain damage varies with this position.

A remaining challenge is that a model's precise position on the spectrum cannot, in general, be directly identified.
We therefore introduce practical proxies that render the analysis tractable within our framework.
First, in the simple setting of \textit{modular addition}, we exploit \textbf{grokking} \citep{nanda2023progress}, which provides a clear binary split between memorizing and generalizing models.
We find that memorizing models suffer minimal retain damage, whereas generalizing models pay a much higher price. 
With generalization realized as a Fourier circuit \citep{nanda2023progress}, we confirm that unlearning disrupts this circuit and thereby degrades retain performance.

Meanwhile, real-world tasks typically demand a combination of memorization and generalization \citep{zhang2023counterfactual,morris2026memorize}.
To capture this setting, we introduce \textit{bucketed modular addition} (BMA), which controls the balance between the two by design.
In particular, BMA consists of two parts: (1) selecting a bucket via modular addition, which rewards generalization, and (2) identifying the index within the bucket, which is randomly assigned and thus requires memorization.
While fixing the total number of answer candidates, adjusting the bucket size (e.g., 4 buckets $\times$ 4 items = 2 buckets $\times$ 8 items = 16 candidates) let us tune how much the task relies on memorization versus generalization.
We quantify this as the \textbf{memorization floor}, the number of bits per answer that must be memorized beyond the shared rule.
In this setup, we reaffirm that configurations with a higher memorization floor (i.e., less generalization) suffer less retain damage, and that this trend is monotonic across the spectrum: each step toward generalization costs more retain damage.

Moreover, we extend our analysis to LLM unlearning, where regularities in natural language play the role of generalization by making part of what must be recalled predictable. 
The sample-specific remainder plays the role of the memorization floor, which we estimate via \citeauthor{morris2026memorize}'s (\citeyear{morris2026memorize}) \textbf{unintended memorization}, i.e., the information a model holds beyond what a reference model predicts. We consider two settings: in verbatim recall, \textit{linguistic regularities} reduce what a model must memorize to reproduce a passage \citep{bengio2003neural,prashanth2025recite}; in factual recall, \textit{semantic regularities} shared across entities narrow the space of plausible answers \citep{chughtai2024summing,liu2025relationspecific}. On the Hubble model suite \citep{wei2025hubblemodelsuiteadvance}, a higher estimated memorization floor again predicts less retain damage in both settings, confirming that the relationship observed in BMA extends to language modeling.

Finally, we discuss two practical implications. First, our findings distinguish two objectives often conflated under unlearning: removing a forget set's contribution to training and suppressing the answers it induced. Although retraining without the forget set constitutes the gold standard for the former \citep{bourtoule2021unlearning}, a retrained model may nonetheless reproduce the targeted answers from the remaining data when generalization accounts for most of them \citep{cooper2025machine,liu2025verbatim}; methods should therefore state explicitly which objective they pursue. Second, the probability mass that unlearning removes from a targeted answer is necessarily reassigned, and its destination proves consequential: redirecting it to answers consistent with the shared rule incurs substantially less retain damage than redirecting it to arbitrary ones. Unlearning objectives should accordingly favor alternatives that preserve the structure underlying generalization.

\begin{figure}
  \centering
  \includegraphics[width=0.95\linewidth]{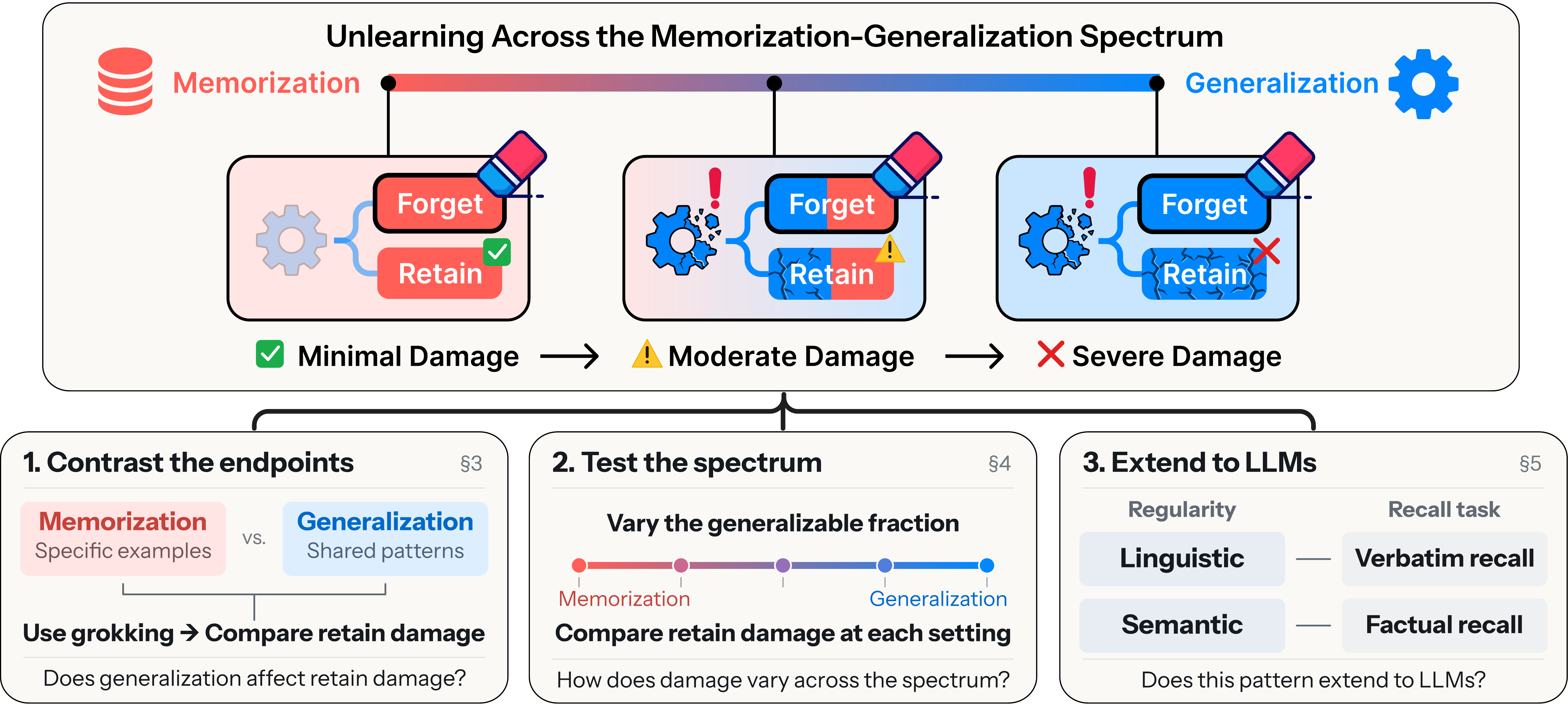}
  \caption{\textbf{Analysis of unlearning across the memorization--generalization spectrum.}
  We argue that retain damage in unlearning grows with a model's reliance on generalization over memorization during training.
  We study this in three steps: contrasting memorizing and generalizing models separated by grokking (\S\ref{sec:grokking}), extending this contrast to a fine-grained analysis across the spectrum, quantified by the memorization floor (\S\ref{sec:bma}), and confirming the relationship in LLM unlearning (\S\ref{sec:llm}).}
  \label{fig:main}
\end{figure}

\section{Background and Related Work}
\label{sec:background}
\subsection{Memorization and Generalization}
\label{sec:background-generalization}
Generalization enables a model to learn patterns across the dataset that transfer to unseen examples from the same distribution \citep{bengio2013representation, liu2022towards}.
Memorization, by contrast, captures what is tied to particular training examples rather than to the underlying distribution.
Mechanistic analyses show that the two correspond to two distinct states of parameters in the model during training \citep{nanda2023progress}.
The same distinction applies to large language models, where memorization corresponds to storing specific knowledge, while generalization refers to linguistic or semantic regularities which the model exploits \citep{prashanth2025recite, chughtai2024summing}.
Recent work by \citet{morris2026memorize} observes the shift between two states when model capacity is saturated, and formalizes and quantifies generalization and memorization through an information-theoretic decomposition.
We use the two terms in a specific sense throughout this paper: \textbf{memorization} denotes learning information tied to individual training samples, whereas \textbf{generalization} denotes learning a shared solution mechanism that applies across examples.

\subsection{Machine Unlearning}
\label{sec:background-unlearning}

Let $\mathcal{D}$ denote the training dataset, $\mathcal{D}_f \subseteq \mathcal{D}$ the \textit{forget set}, and $\mathcal{D}_r = \mathcal{D} \setminus \mathcal{D}_f$ the \textit{retain set}. 
The goal of machine unlearning is to remove information learned from $\mathcal{D}_f$ while preserving performance on $\mathcal{D}_r$ \citep{cao2015unlearning,ginart2019making}. Retraining from scratch without $\mathcal{D}_f$ provides an ideal gold standard for removing its influence \citep{bourtoule2021unlearning}. 
However, removing information from $\mathcal{D}_f$ often degrades performance on $\mathcal{D}_r$ \citep{maini2024tofu,shi2025muse}. 
Therefore, effective unlearning aims to minimize the trade-off between forgetting and utility preservation, which remains a central objective in machine unlearning.

Prior studies have examined unlearning difficulty through data entanglement but have not directly connected it to generalization and memorization.
For instance, \citet{whatmakesunlearninghard} show that unlearning becomes more challenging when individual forget examples receive higher post-training probabilities or are more entangled with the retain set in embedding space.
Similar difficulty arises at the pre-training scale, where knowledge appearing more frequently in the corpus is harder to selectively remove \citep{krishnan2025not}.
More recent studies show an analysis of a specific unlearning method by distinguishing generation by learned rules from those recalled by memorization \citep{zhou2026the}.
This perspective is also adopted by work that separates components redundant with the retain set from target-data information, enabling better retain preservation \citep{wei2026rethinkingfederatedunlearninglens}.
However, the relationship between unlearning, generalization, and memorization remains largely unexplored. 
This work addresses that gap by investigating how the latter two shape the former.

\section{Memorization vs. Generalization: A Binary View}
\label{sec:grokking}

To systematically study retain damage in unlearning, we look back at how models learn. Specifically, models can develop generalizable solutions by exploiting common patterns, rather than simply memorizing individual training examples \citep{bengio2013representation,liu2022towards}.
We hypothesize that unlearning causes retain damage when it disrupts such a mechanism shared by $\mathcal{D}_f$ and $\mathcal{D}_r$.
Testing this hypothesis requires models that fit the same data equally well but differ in whether they memorize or generalize, which is difficult to determine in practice.
We address this challenge through a controlled toy setting that exhibits \textbf{grokking}: a sharp rise of test accuracy long after a model has fit its training data \citep{power2022grokking}.
In this setting, mechanistic tracing of the transition allows us to compare memorization and generalization and examine how they shape retain damage in unlearning.

\subsection{Grokking as a Binary Lens in Modular Addition}
\label{sec:modular-setting}

\paragraph{A toy model of grokking.}

\citet{nanda2023progress} demonstrate that grokking occurs in a one-layer transformer trained to predict $y=(a+b)\bmod p$ for $p=113$. 
Their mechanistic analysis traces three phases in the model's solution strategy. 
The first, \textit{memorization}, has the model fit the training examples while generalizing poorly. 
In the second, \textit{circuit formation}, a generalizing Fourier circuit develops while memorizing components still support predictions, before test performance takes off.
Finally, in \textit{cleanup}, weight decay encourages the model to shed the memorizing components in favor of the more efficient circuit, and test performance improves sharply.
The resulting grokked (i.e., generalized) model solves training and unseen examples through the same algorithm.

\paragraph{What remains to unlearn after cleanup?}
This transition yields two models that solve the training set equally well but differ in strategy: the memorizing model stores each example individually, whereas the grokked (generalized) model relies entirely on the shared circuit. 
We hypothesize that unlearning the latter must damage this circuit, on which retained examples also depend, and thus incurs greater retain damage.

\subsection{Generalization Is a Key Cause of Retain Damage}
\label{sec:strategy-damage}

\paragraph{Experimental setup.}
To test this prediction, we follow \citet{nanda2023progress} and train the toy transformer on 30\% of all input pairs, holding out the remaining pairs for evaluation. 
To our knowledge, this work constitutes the first attempt to interpret unlearning via the lens of grokking.
We randomly select 10\% of the training set as $\mathcal{D}_f$ and use the rest as $\mathcal{D}_r$, using the same forget and retain examples at every checkpoint within each $\mathcal{D}_f$ selection seed. 
We select four representative checkpoints from a single training run spanning memorization, circuit formation, cleanup, and the fully grokked regime. 
All have 100\% training accuracy, while their test accuracy ranges from poor generalization to 100\%. 
We apply RL \citep{fan2024salun} and GD \citep{largelanguagemodelunlearningGD} until forget accuracy first reaches the chance level ($\tau=1/113$) or below.
We repeat this comparison across five $\mathcal{D}_f$ selection seeds, keeping the training run and train/test split fixed. 
Implementation details are in Appendix~\ref{app:grokking-config}.

\paragraph{Results.}

\begin{wraptable}{r}{0.48\textwidth}
  \vspace{-1.75\intextsep}
  \setlength{\belowcaptionskip}{5pt}
  \centering
  \caption{\textbf{Retain accuracy (\%).} Mean $\pm$ 95\% CI across five $\mathcal{D}_f$ selection
  seeds when forget accuracy first falls to a threshold $\tau$ or below.}
  \label{tab:grokking-chance}
  \small
  \setlength{\tabcolsep}{3pt}
  \begin{tabular}{@{}lrrr@{}}
    \toprule
    Phase & Epoch & RL & GD \\
    \midrule
    Memorization & 1,000 & 59.5\,\textpm{}\,3.3 & 23.0\,\textpm{}\,6.5 \\
    Circuit formation & 3,800 & 52.4\,\textpm{}\,7.3 & 19.3\,\textpm{}\,5.3 \\
    Cleanup & 10,000 & 14.2\,\textpm{}\,1.5 & 8.7\,\textpm{}\,3.2 \\
    Grokked & 25,300 & 8.1\,\textpm{}\,0.7 & 3.3\,\textpm{}\,1.0 \\
    \bottomrule
  \end{tabular}
  \par\vspace{-0.7\intextsep}
\end{wraptable}

The models respond differently to unlearning despite solving every training example correctly beforehand.
At chance-level forget accuracy, mean retain accuracy under RL is 59.5\% for the memorizing model and 8.1\% for the grokked model; GD yields corresponding accuracies of 23.0\% and 3.3\% (Table~\ref{tab:grokking-chance}).
Models sampled during circuit formation and cleanup fall between these two endpoints. Figure~\ref{fig:grokking-trajectories} (top row) follows the evolution of forget, retain, and test accuracy over unlearning steps, showing retain and test accuracy falling together in the grokked model.

These results support our hypothesis: the greater retain damage in the generalized model is consistent with unlearning acting on a shared solution mechanism when little sample-specific memorization remains. 
As this mechanism has been identified as a Fourier circuit in the toy model, we next test the explanation mechanistically by examining whether unlearning damages the circuit itself.

\subsection{Fourier Circuit as a Concrete Example of Generalization}

\label{sec:mechanistic-attribution}
\begin{figure}[t]
  \centering
  \includegraphics[width=\linewidth]{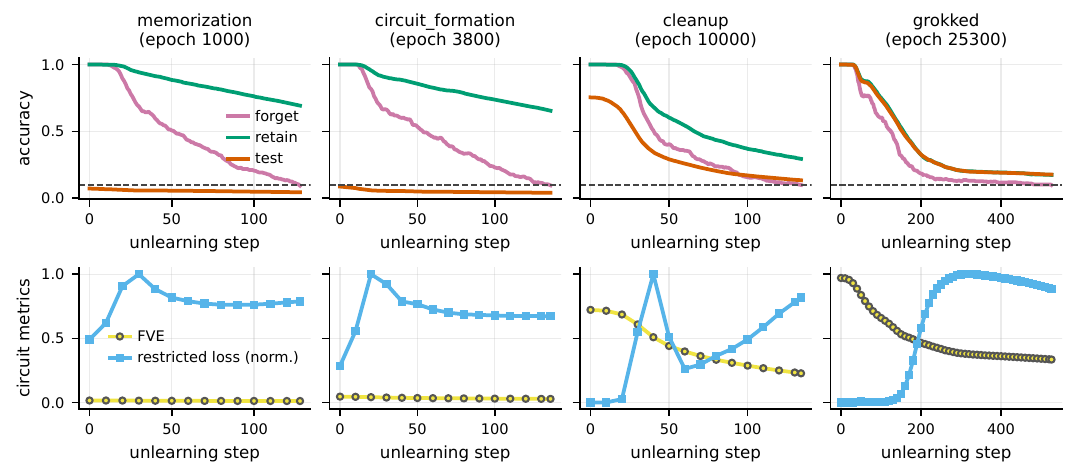}
  \caption{\textbf{Learning strategy changes the retain damage of unlearning.} Unlearning trajectories under random labeling (RL) at four checkpoints from memorization to grokking ($\mathcal{D}_f$ selection seed 1). Top: forget accuracy falls much faster than retain accuracy in the memorizing model, whereas both decline together in the generalized model, incurring greater retain damage. Bottom: the Fourier circuit contributes little in the memorizing model; in the generalized model, falling FVE and rising restricted loss reveal its degradation. Restricted loss is normalized per trajectory.}
  \label{fig:grokking-trajectories}
\end{figure}

\paragraph{The Fourier circuit and key frequencies.}
\citet{nanda2023progress} show that the grokked toy transformer's logit vector is well approximated by a weighted sum of Fourier components at a small set of frequencies, termed \emph{key frequencies}. 
At each frequency $\omega_k=2\pi k/p$, the circuit combines sine and cosine representations of the inputs through angle-addition identities to produce a logit contribution proportional to $\cos(\omega_k(a+b-c))$ for each candidate answer $c$.
At $c=(a+b)\bmod p$, these components align and interfere constructively, giving the correct answer a high logit. 
At incorrect candidates, their differing phases lead to partial cancellation and lower logits. 
In short, these key frequency components implement the shared solution mechanism for modular addition.

\paragraph{Tracing circuit damage.}
We fix the key frequencies identified in the fully trained model and isolate the corresponding circuit components in the logits. 
We track two complementary measurements:
\begin{enumerate}
  \setlength{\itemsep}{2pt}
  \setlength{\parskip}{0pt}
  \item \textbf{FVE} (Fraction of Variance Explained) is the fraction of variance in the logits explained by the key frequency components, indicating how much the Fourier circuit contributes to generating the model's output.
  \item \textbf{Restricted loss} \citep{nanda2023progress} is the test cross-entropy loss computed using logits reconstructed from the key frequency components.
  It measures how well the Fourier circuit can solve the task on its own, with lower values indicating better performance.
\end{enumerate}

In the memorizing model, FVE remains low and restricted loss remains high before and after unlearning, consistent with its limited reliance on the Fourier circuit (Figure~\ref{fig:grokking-trajectories}, bottom row). 
In contrast, unlearning the generalized, grokked model reduces FVE from 0.97 to 0.33 and increases restricted loss from near zero to 2.44, indicating degradation of the Fourier circuit. 
These changes accompany the decline in forget accuracy and the collapse in retain accuracy, linking the observed retain damage to impairment of the shared solution mechanism.
We further test whether circuit damage causally contributes to retain damage through an intervention (Appendix~\ref{app:causal-intervention}).

\section{Fine-Grained Analysis of Unlearning across the Spectrum}
\label{sec:bma}

Section~\ref{sec:grokking} provides a binary comparison between memorizing and generalizing models on a fully algorithmic task.
In this setting, every component needed to determine the answer is generalizable: once the shared rule is learned, no sample-specific information is required.
This leaves open how unlearning behaves when generalization accounts for only part of an answer.
For example, when completing ``Beats Music is owned by,'' general linguistic patterns can narrow the candidates to company names, but selecting ``Apple'' requires memorization of the specific ownership relation.
To study this, we introduce \emph{bucketed modular addition} (BMA), which constrains by design how much of an answer can be determined by a shared rule and how much must be memorized individually.
By varying this balance, we explore intermediate positions along the memorization--generalization spectrum and examine whether greater scope for generalization leads to greater retain damage.

\subsection{Bucketed Modular Addition}
\label{sec:bma-task}

\paragraph{Task design.}

To instantiate this spectrum, we introduce \emph{bucketed modular addition} (BMA) to control how much a shared rule determines while keeping the answer space fixed at 256 classes.
For inputs $a,b\in\{0,\ldots,255\}$, and the bucket size $M$, the correct answer is
{
\small
\begin{equation}
  y(a,b)=M\bigl((a+b)\bmod p\bigr)+u_{a,b},
  \qquad pM=256,
  \label{eq:bma-task}
\end{equation}
}
where $u_{a,b}$ is sampled independently and uniformly from
$\{0,\ldots,M-1\}$ for each input pair, then fixed throughout experiments.
The modular rule determines which bucket to select, narrowing the 256 possible answers to the $M$ members of that bucket.
The random assignment of the correct member within each bucket does not generalize across input pairs and therefore requires memorization.

Varying $M$ therefore controls how much uncertainty the shared rule can resolve and how much remains for memorization.
At the minimum bucket size ($M=1$), the task reduces to modular addition and the shared solution determines every answer.
At the maximum bucket size ($M=256$), all answers belong to a single bucket and the labels are entirely random, leaving no room for generalization.
Intermediate settings occupy different positions on the spectrum, balancing bucket selection through the modular rule with memorization of the correct member within that bucket.

\paragraph{The memorization floor.}
We define the \textbf{memorization floor} as the minimum amount of sample-specific information needed to capture non-generalizable information, even after a model has learned the shared solution mechanism.
In BMA, the modular rule selects the correct bucket but provides no further information to distinguish among its $M$ equally likely members.
The remaining information needed to identify the correct member must be memorized individually, so the memorization floor is computed as $\log_2{M}$ bits per example.
With $pM=256$ fixed, specifying each answer requires 8 bits of information in total, and adjusting $M\in\{1,4,16,64,256\}$ lets us control how much of this information must be memorized individually---the memorization floor---from none to all 8 bits.

\subsection{The Memorization Floor Predicts Retain Damage}
\label{sec:excess-prediction}

\paragraph{Experimental setup.}
Following the experimental setup in Section~\ref{sec:strategy-damage}, we train a one-layer transformer for each $M\in\{1,4,16,64,256\}$ on the same 30\% of input pairs for 30,000 epochs.
We randomly select 1\%, 5\%, or 10\% of the training set as $\mathcal{D}_f$, use the remaining examples as $\mathcal{D}_r$, and repeat each comparison across three $\mathcal{D}_f$ selection seeds.
We apply GD, RL, DPO \citep{rafailov2023direct}, and ME \citep{yuan2025a}, using the same forget/retain split across bucket sizes and methods within each seed and forget fraction.
To compare retain damage at matched levels of forgetting, we denote the mean correct-answer probabilities on $\mathcal{D}_f$ and $\mathcal{D}_r$ by $P_f$ and $P_r$, respectively, and measure $1-P_r$ at $P_f=0.1$.
Method and implementation details are provided in Appendix~\ref{app:bma-config}.

\begin{figure}[t]
  \centering
  \setlength{\abovecaptionskip}{4pt}
  \includegraphics[width=0.95\linewidth]{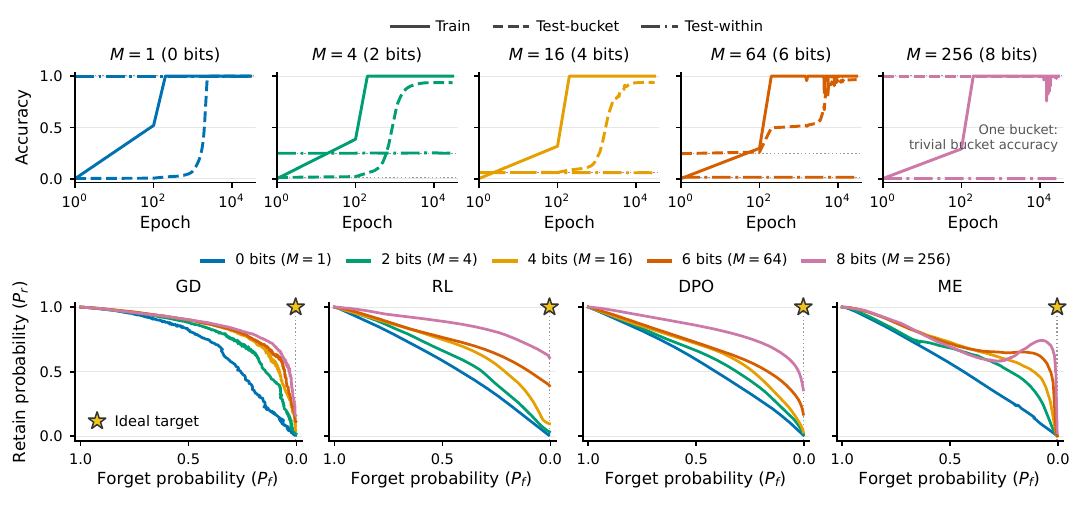}
  \caption{\textbf{Grokking and the forget--retain trade-off in BMA.}
  Top: For $M\leq64$, test bucket accuracy (dashed) rises long after training
  exact accuracy (solid) saturates, revealing grokking in bucket selection. Dotted lines mark chance levels.
  Bottom: Forget--retain trade-offs in $P_f$
  and $P_r$ during unlearning (5\% forget set, seed 17). With larger memorization floor, the
  trade-off curves generally lie closer to the starred ideal target:
  $P_f=1/256$ and $P_r=1$.}
  \label{fig:bma-trajectories}
\end{figure}

\paragraph{Learning dynamics in BMA.}
We track two test metrics throughout training.
\emph{Bucket accuracy} ($A_{\mathrm{bucket}}$) measures how often the predicted answer falls in the correct bucket, indicating how well models have learned the modular rule.
\emph{Within-bucket accuracy} ($A_{\mathrm{within}}$) measures how often the highest-scoring answer within the correct bucket is correct, indicating how well models distinguish among its members.
For $M\in\{1,4,16,64\}$, $A_{\mathrm{bucket}}$ approaches 1 well after training accuracy reaches 100\%, revealing grokking in bucket selection, while $A_{\mathrm{within}}$ remains near its chance level of $1/M$ (Figure~\ref{fig:bma-trajectories}, top row).\footnote{At $M=256$, bucket accuracy is trivially 100\% because there is only one bucket.}
These results show that, as designed, generalization resolves bucket selection but leaves the selection of the correct member within each bucket to individual memorization.
We use these trained models at different positions along the memorization--generalization spectrum to examine how retain damage varies with their required reliance on individual memorization.

\paragraph{Unlearning results.}
Figure~\ref{fig:bma-trajectories} (bottom) shows that higher memorization floors generally correspond to higher $P_r$ at comparable $P_f$.
Figure~\ref{fig:bma-excess} summarizes retain damage $1-P_r$ at $P_f=0.1$, averaged over three seeds.
Across settings, mean retain damage generally decreases as the memorization floor increases.
These results extend the memorization--generalization contrast in Section~\ref{sec:grokking} to intermediate positions along the spectrum.
Tasks requiring more individual memorization incur less retain damage, whereas those allowing generalization to account for more of the answer incur greater damage at matched levels of forgetting.
This pattern supports our hypothesis that a model's reliance on memorization versus generalization shapes the retain damage of unlearning.

\section{The Spectrum in Practice: LLM Unlearning}
\label{sec:llm}

In BMA, the regularity that enables generalization is explicitly built into the task: the modular-addition rule.
To examine whether our findings extend beyond this controlled setting, we turn to language modeling, where regularities in natural language may likewise reduce the need for sample-specific memorization.
We consider two settings: verbatim recall, where \emph{linguistic regularities} can make passages more predictable \citep{bengio2003neural,prashanth2025recite}, and factual recall, where \emph{semantic regularities} can make answers more predictable across entities \citep{chughtai2024summing,liu2025relationspecific}.
Using these two settings, we examine whether the relationship between position on the spectrum and retain damage observed in BMA also applies to LLM unlearning.

\paragraph{Estimating the memorization floor in LLMs.}
In BMA, the memorization floor is determined by task design.
In LLMs, however, the shared regularities and the sample-specific information required beyond them are not explicitly specified.
We therefore use \textbf{unintended memorization} \citep{morris2026memorize} as a practical proxy, estimating how much sample-specific information a target model holds beyond reference model.
Specifically, we estimate this information using the log-probability gain $\log_2 p_{\mathrm{target}}(x)-\log_2 p_{\mathrm{ref}}(x)$, where the target has been trained on $x$ and the reference captures shared regularities without having been trained on $x$.
This estimate coincides with the memorization floor in BMA under an idealized target--reference comparison.
Consider a reference model that knows the correct bucket but not the correct member, and a target model that additionally knows the correct member.
If the reference distributes probability uniformly over the $M$ members of the correct bucket and the target assigns probability 1 to the correct answer, their log-probability difference is $\log_2 1-\log_2(1/M)=\log_2 M$, recovering the memorization floor specified by the task design.

\paragraph{The Hubble model suite.}
Hubble \citep{wei2025hubblemodelsuiteadvance} provides paired models suited to this comparison.
Its \emph{standard model} is pretrained on a base corpus filtered to exclude texts designated for controlled insertion, while its \emph{perturbed model} follows the same pretraining procedure with those texts included.
The standard model provides a reference for what is predictable without training on the inserted texts, allowing us to estimate sample-specific memorization as $\log_2 p_{\mathrm{pert}}(x)-\log_2 p_{\mathrm{std}}(x)$.
We use this estimate as a proxy for the memorization floor.
We study Hubble-8B models pretrained on 100B and 500B tokens, with Hubble-1B results reported in Appendices~\ref{app:verbatim-recall-unlearning} and~\ref{app:factual-recall-unlearning}.

\subsection{Linguistic Regularities in Verbatim Recall}
\label{sec:verbatim}

\paragraph{Verbatim recall.}
We study verbatim recall using book passages from Gutenberg Popular \citep{gutenberg}, one of the datasets used for Hubble's controlled text insertions.
For a passage $x=(x_1,\ldots,x_T)$, we evaluate recall under teacher forcing, predicting each token $x_t$ given its ground-truth prefix.
We summarize recall performance using the geometric mean of the correct-token probabilities \citep{maini2024tofu}:
{
\small
\begin{equation}
  \bar{P}(x)=\exp\left(\frac{1}{T}\sum_{t=1}^{T}
  \log p(x_t\mid \texttt{<BOS>},x_{<t})\right).
  \label{eq:verbatim-prob}
\end{equation}
}
For each passage, we estimate the memorization floor by summing the token-level estimates over all $T$ tokens, yielding $\sum_{t=1}^{T}[\log_2 p_{\mathrm{pert}}(x_t\mid \texttt{<BOS>},x_{<t})-\log_2 p_{\mathrm{std}}(x_t\mid \texttt{<BOS>},x_{<t})]$ bits.

\paragraph{Experimental setup.}
We use the $256\times$ subset of Gutenberg Popular, comprising 80 passages.
For each model, we rank these passages by their estimated memorization floor and assign the top and bottom 30\% to the \emph{High} and \emph{Low} forget sets, respectively.
We unlearn each forget set using GA \citep{jang-etal-2023-knowledge}, GD, NPO \citep{zhang2024npo}, and CEU \citep{yang2025ceu}.
At $\bar{P}_f=0.1$, we measure retain damage as the decrease in $\bar{P}_r$ on three groups: retained passages from the same subset (\emph{ID}), inserted Wikipedia passages (\emph{OOD}), and non-inserted passages from both corpora (\emph{unseen}).

\paragraph{Linguistic regularities and the memorization floor.}
We examine whether linguistic regularities in a passage are reflected in the memorization floor estimated from the models.
We measure lexical rarity using TF-IDF, with higher values indicating less frequent vocabulary.
Table~\ref{tab:verbatim-examples} illustrates the correspondence: the High passage contains many rare words and has a higher TF-IDF, whereas the Low passage consists largely of common vocabulary and has a lower TF-IDF.
Across all four models, TF-IDF correlates positively with the estimated memorization floor (Figure~\ref{fig:verbatim-tfidf}).
This association suggests that models exploit linguistic regularities to reduce what they must memorize individually, while passages containing rarer vocabulary require more individual memorization. 

\begin{table}[t]
  \centering
  \setlength{\belowcaptionskip}{6pt}
  \caption{\textbf{Rarer vocabulary, higher memorization floor.} One example passage from each forget set; words occurring five or fewer times in inserted Gutenberg Popular passages are highlighted.}
  \label{tab:verbatim-examples}
  \scriptsize
  \setlength{\tabcolsep}{4pt}
  \begin{tabular}{@{}lp{0.65\textwidth}rr@{}}
    \toprule
    Memorization Floor& Passage & TF-IDF & Bits \\
    \midrule
    High & \ldots\, those of \hl{Humelbergius}, the \hl{editor}-physician of \hl{Z\mbox{\"u}rich}, will be \hl{enjoyed} and read with \hl{profit} by every \hl{antiquary}. The \hl{labors} of \hl{Bernhold} and \hl{Schuch} are \hl{meritorious}\,\ldots & 7.29 & 1268 \\[2pt]
    \midrule
    Low & \ldots\,just as \hl{obvious} as the principles taken for granted. For no very good reason, three of these principles have been \hl{singled} out by \hl{tradition} under the name of `Laws of Thought' \,\ldots & 6.68 & 614 \\[2pt]
    \bottomrule
  \end{tabular}
  \par\vspace{-0.5\intextsep}
\end{table}

\begin{figure}[t]
  \centering
  \setlength{\abovecaptionskip}{0pt}
  \includegraphics[width=\linewidth]{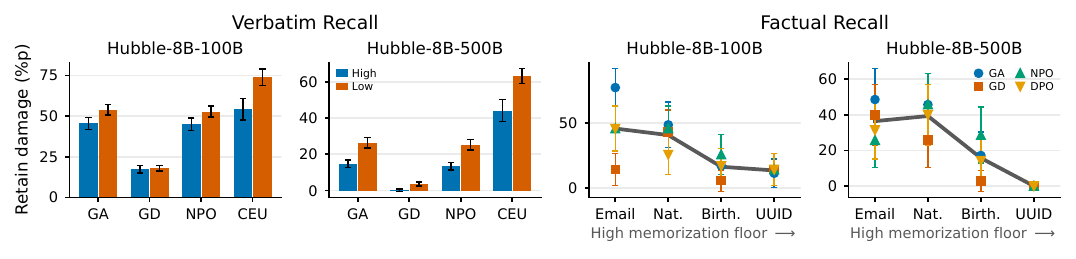}
  \caption{\textbf{A higher memorization floor leaves less retain damage.}
  (Left)
  Damage on ID $\pm$ 95\% CI after
  unlearning the High and Low forget sets.
  (Right)
  In-relation damage $\pm$ 95\% CI for each target relation for each method, ordered
  by the relation's memorization floor.
  Heavy line represents the means over the four methods.
  Nat.\ and Birth.\ abbreviate nationality and birthplace respectively.
  }
  \label{fig:verbatim-damage}
  \label{fig:yago-damage}
\end{figure}

\paragraph{Retain damage analysis.}
Comparing the High and Low forget sets reveals two patterns (Figure~\ref{fig:verbatim-damage}).
(1) At the same level of forgetting, unlearning the High forget set generally causes less retain damage than unlearning the Low forget set across settings.
This suggests that suppressing passages with a lower memorization floor places greater pressure on shared linguistic mechanisms, extending the relationship observed in BMA to verbatim recall.
(2) Damage is generally greater on ID than on OOD passages.
This further supports our account that retained predictions are more vulnerable when they rely on linguistic regularities shared with the forget set.

\subsection{Semantic Regularities in Factual Recall}
\label{sec:factual}

\paragraph{Factual recall.}
We study factual recall using synthetic YAGO biographies \citep{Tanon2020YAGO4A} from Hubble's controlled text insertions, considering four relations: nationality, birthplace, email, and UUID.
Given an entity name $s$ and a relation $r$, the task is to identify the correct attribute value $a$ from a fixed candidate set $\mathcal{A}$ containing the correct answer and 15 distractors from the same relation.
We measure recall performance by accuracy, counting a prediction as correct when the correct answer has the highest conditional likelihood among the 16 candidates.
To estimate the memorization floor, we normalize the answer-string likelihoods over the candidate set:
{
\small
\begin{equation}
  P_m^{16}(a\mid s,r,\mathcal{A})=
  \frac{p_m(a\mid s,r)}{\sum_{b\in\mathcal{A}}p_m(b\mid s,r)}.
  \label{eq:p16}
\end{equation}
}
We estimate each fact's memorization floor as $\log_2 P_{\mathrm{pert}}^{16}(a\mid s,r,\mathcal{A})-\log_2 P_{\mathrm{std}}^{16}(a\mid s,r,\mathcal{A})$ bits.

\paragraph{Experimental setup.}
For each model and relation, we select 50 correctly recalled facts from the $64\times$ subset of the YAGO biographies and randomly assign 15 (30\%) to $\mathcal{D}_f$, leaving the remaining 35 as $\mathcal{D}_r$.
For each of four relations, we unlearn the corresponding forget set using GA, GD, NPO, and DPO.
When forget accuracy falls to 10\%, we measure retain damage as the decrease in accuracy on the retained facts of the target relation (\emph{in-relation}) and those of the other relations (\emph{out-relation}).

\paragraph{Semantic regularities and the memorization floor.}
We examine whether semantic regularities shared across entities are reflected in the memorization floor estimated from the models.
\begin{wraptable}{r}{0.49\textwidth}
  \vspace{-\intextsep}
  \setlength{\abovecaptionskip}{0pt}
  \setlength{\belowcaptionskip}{6pt}
  \centering
  \caption{\textbf{Predictable attributes have a lower memorization floor.} Hubble-8B-500B examples; memorization floor in bits.}
  \label{tab:yago-examples}
  \scriptsize
  \setlength{\tabcolsep}{3pt}
  \begin{tabular}{@{}llr@{}}
    \toprule
    Relation & Subject $\,\rightarrow\,$ Attribute & Mem. Floor \\
    \midrule
    Email & Maurice Clauss $\,\rightarrow\,$ \texttt{clauss@gmail.com} & 0.0002 \\
    \addlinespace[3pt]
    Nat. & Ricardo da Sousa$\,\rightarrow\,$ \texttt{Brazil}  & 0.2013 \\
    \addlinespace[3pt]
    UUID & Tony Nieminen \mbox{$\, \rightarrow\,$ \texttt{766e00}\ldots\texttt{8b12ce}} & 5.6064 \\
    \bottomrule
  \end{tabular}
  \par\vspace{-0.8\intextsep}
\end{wraptable}

Table~\ref{tab:yago-examples} illustrates the correspondence.
Email addresses in YAGO follow name-based formats, such as \texttt{clauss@gmail.com} for Maurice Clauss, making them largely predictable from names.
Naming conventions also provide partial clues to nationality, as in Ricardo da Sousa's Brazilian nationality, while randomly assigned UUIDs lack such predictive structure.
In both models, the mean memorization floor per relation is lower for email and nationality (0.43--1.61 bits) and higher for birthplace and UUID (2.30--5.70 bits; Table~\ref{tab:yago-prompts}).
This pattern suggests that models exploit semantic regularities to reduce what they must memorize individually, while less predictable attributes require more memorization.

\paragraph{Retain damage analysis.}
Comparing retain damage across relations reveals similar patterns as seen in verbatim recall.
(1) At comparable levels of forgetting, unlearning email and nationality, which have lower memorization floors, generally causes more in-relation retain damage than unlearning birthplace and UUID, which have higher memorization floors (Figure~\ref{fig:yago-damage}).
These results suggest that suppressing facts with a lower memorization floor places greater pressure on shared semantic mechanisms, extending the relationship observed in BMA 
to factual recall.
(2) The gap between in-relation and out-relation retain damage is generally larger for relations with lower memorization floors (Table~\ref{tab:yago-method-mean}).
This is consistent with our hypothesis that facts with lower memorization floors rely more on shared mechanisms and therefore suffer greater retain damage in unlearning.

\section{Discussion}
\label{sec:discussion}

\paragraph{The retraining standard and targeted behavior suppression.}
Our findings reveal that two goals grouped under unlearning do not always coincide:
removing training contributions and suppressing targeted behavior.
Treating these goals as equivalent assumes that the targeted behavior depends
exclusively on learning from $\mathcal{D}_f$. This premise can fail when the memorization floor is small and shared solution mechanisms account for most of the
prediction. In the case of modular addition, where memorization floor is zero, the rule fully determines
the answer and can be independently learned from examples outside $\mathcal{D}_f$. A model
retrained from scratch without $\mathcal{D}_f$ may therefore still produce the
targeted outputs. Thus, retraining without the forget set is a gold standard for
removing its training contribution, but does not always achieve the behavioral
goal of suppressing outputs.

\paragraph{Within-bucket redistribution reduces retain damage.}
In BMA, we compare RL with incorrect targets drawn from the full answer space or
restricted to the correct bucket. At 10\% forget accuracy, within-bucket targets
preserve substantially higher retain exact accuracy and forget bucket accuracy
(Table~\ref{tab:bma-redistribution}). This intervention suppresses
individual answers while keeping replacement targets consistent with the modular
rule, reducing pressure to damage the shared solution mechanism. Selective forgetting
thus depends not only on how much probability is removed from a target, but also on
where it is reassigned. Appendix~\ref{app:redistribution} provides experimental details
and considers how this principle could guide existing methods and LLM unlearning.

\section{Conclusion}
\label{sec:conclusion}
We study how the way models learn shapes retain damage in unlearning.
Across the memorization--generalization spectrum, greater reliance on shared solution mechanisms generally leads to greater retain damage.
In modular addition, generalizing models suffer more damage than memorizing models as unlearning disrupts the shared circuit.
BMA extends this comparison to intermediate positions on the spectrum, and LLM experiments reveal a similar pattern for linguistic and semantic regularities.
Building on the implications of our findings, we discuss the distinction between contribution removal and output suppression, and the benefits of preserving shared structure when redistributing probability.
Overall, our work deepens the understanding of why retain damage occurs in unlearning by linking it to generalization in learning, a connection that has yet to be fully explored.

\subsection*{AI use statement}
Among the subtasks for which disclosure is required, we used them throughout the research process,
and we describe here how they were used and how we verified their output.
In this work, we used generative AI tools to help develop theoretical models or conceptual frameworks, design and provide feedback on research methodology and experiments, implement methods, help with translation and interpret results.
We have not used generative AI tools to propose or refine hypothesis, or to support qualitative and thematic data analysis; and generating synthetic data sets, formulating mathematical claims, providing critical ingredients for proving mathematical claims, assisting in the writing of proofs, or cleaning and reformatting dataset is not applicable to this work.
Additionally, we used generative AI tools for creating and modifying scientific figures or images, creating and editing software code, drafting and editing parts of this paper for readability, suggesting its structure, and identifying and summarizing related literature.

We reviewed all AI-assisted work.
Specifically, the terms and framing suggested by AI was never used directly, such suggestions served as inspiration from which we derived our own framing, and the statistics for our experiments were checked by multiple authors.
The authors significantly revised the text drafted or translated with assistance and the literature summarized by AI was reviewed in full detail, verifying the claim in the cited paper.
Code written with AI assistance, both the implementations of the experiments and script summarizing the experimental result were read and tested by the authors.
Figures and images were obtained through generated python script with raw experimental logs as inpu lt to eliminate hallucination.
We take responsibility for final content of this work, including text, claims, or artifacts produced with the aid of generative AI.

\bibliography{iclr2027_conference}
\bibliographystyle{iclr2027_conference}

\clearpage

\appendix
\section{Unlearning Methods}
\label{app:unlearning-methods}

This section defines the objectives used in our experiments.
All objectives below are minimized; learning rates, retain weights, and comparison criteria are specified in the corresponding experimental sections.
Let $\mathcal{L}_{\mathrm{CE}}(\mathcal{S};\theta)$ denote mean cross-entropy on a set $\mathcal{S}$, using the original labels or tokens.
For LLMs, cross-entropy is averaged over the scored tokens within each example and then over examples: the scored span is the passage for verbatim recall and the answer for factual recall.
Only GD includes a retain-loss term in the implementations considered here.

\paragraph{Gradient ascent (GA).}
GA increases cross-entropy on the forget examples \citep{jang-etal-2023-knowledge} by minimizing
\begin{equation}
  \mathcal{L}_{\mathrm{GA}}=-\mathcal{L}_{\mathrm{CE}}(\mathcal{D}_f;\theta).
\end{equation}

\paragraph{Gradient difference (GD).}
GD combines forget-loss ascent with retain-loss descent \citep{largelanguagemodelunlearningGD,dorna2026openunlearning}:
\begin{equation}
  \mathcal{L}_{\mathrm{GD}}=-\mathcal{L}_{\mathrm{CE}}(\mathcal{D}_f;\theta)
  +\lambda\mathcal{L}_{\mathrm{CE}}(\mathcal{A};\theta),
\end{equation}
where $\mathcal{A}$ is the retain anchor set used by the experiment.
The experimental sections specify its membership and overlap with the evaluation groups.

\paragraph{Random labeling (RL).}
RL minimizes cross-entropy toward fixed incorrect labels or answers \citep{fan2024salun}:
\begin{equation}
  \mathcal{L}_{\mathrm{RL}}=\mathcal{L}_{\mathrm{CE}}(\widetilde{\mathcal{D}}_f;\theta).
\end{equation}
Here $\widetilde{\mathcal{D}}_f$ replaces each original answer with an incorrect alternative, sampled once and held fixed throughout unlearning and across compared checkpoints.
For modular addition, alternatives are sampled uniformly from the $p-1$ incorrect answer labels; BMA uses the other output classes, and factual recall uses incorrect candidate values.
Appendix~\ref{app:bma-redistribution} compares unrestricted and within-bucket alternatives in a separate BMA intervention.

\paragraph{Direct preference optimization (DPO).}
DPO \citep{rafailov2023direct} prefers a fixed incorrect answer $\tilde y$ over the original answer $y$ relative to the frozen pre-unlearning model $\theta_0$.
Writing $s_\theta(x,y)$ for the answer log-likelihood and $\Delta_\theta(x,y)=s_\theta(x,y)-s_{\theta_0}(x,y)$, its objective is
\begin{equation}
  \mathcal{L}_{\mathrm{DPO}}=-\mathbb{E}_{(x,y)\in\mathcal{D}_f}
  \log\sigma\!\left(\beta[\Delta_\theta(x,\tilde y)-\Delta_\theta(x,y)]\right).
\end{equation}
For BMA, $s_\theta$ is the single-label log probability; for factual recall it is the sum of log probabilities over the answer tokens.

\paragraph{Negative preference optimization (NPO).}
NPO \citep{zhang2024npo} treats the original answer as dispreferred without specifying a preferred alternative:
\begin{equation}
  \mathcal{L}_{\mathrm{NPO}}=-\frac{2}{\beta}\mathbb{E}_{(x,y)\in\mathcal{D}_f}
  \log\sigma\!\left(-\beta\Delta_\theta(x,y)\right).
\end{equation}
In verbatim recall, $s_\theta$ is the token-averaged passage log-likelihood; in factual recall it is the summed answer log-likelihood.
The frozen pre-unlearning model used by NPO and DPO is distinct from the Hubble standard model used to estimate memorization floor.

\paragraph{Maximum entropy (ME).}
ME \citep{yuan2025a} maximizes entropy over the full BMA answer space $\mathcal{Y}$:
\begin{equation}
  \mathcal{L}_{\mathrm{ME}}=\mathbb{E}_{x\in\mathcal{D}_f}
  \left[\log|\mathcal{Y}|-H\bigl(p_\theta(\cdot\mid x)\bigr)\right].
\end{equation}
The constant leaves the gradient unchanged and expresses the objective as KL divergence to the uniform answer distribution.

\paragraph{Cross-entropy unlearning (CEU).}
CEU \citep{yang2025ceu} replaces the original token target with the model's distribution over the remaining vocabulary.
Let $\tilde q_t$ be the softmax of $z_t$ after setting the original token's logit to $-\infty$, with gradients stopped through $\tilde q_t$.
We minimize
\begin{equation}
  \mathcal{L}_{\mathrm{CEU}}=-\mathbb{E}_{x\in\mathcal{D}_f}\frac{1}{T}\sum_{t=1}^{T}
  \sum_v\tilde q_{t,v}\log p_\theta(v\mid x_{<t}).
\end{equation}
The target distribution is recomputed from the current logits at every update.

\FloatBarrier
\section{Grokking Experiments}
\label{app:grokking}

\subsection{Configuration}
\label{app:grokking-config}
We train a one-layer ReLU transformer on modular addition following \citet{nanda2023progress}.
Tables~\ref{tab:grokking-architecture}, \ref{tab:grokking-training-setup}, and~\ref{tab:grokking-unlearning-setup} summarize the model architecture, training configuration, and unlearning configuration, respectively.
The four representative checkpoints span memorization, circuit formation, cleanup, and the grokked regime, and all have 100\% training accuracy.

\begin{table}[htbp]
  \centering
  \setlength{\belowcaptionskip}{6pt}
  \caption{\textbf{Model architecture for modular addition.}}
  \label{tab:grokking-architecture}
  \small
  \renewcommand{\arraystretch}{1.12}
  \begin{tabular}{@{}p{0.36\linewidth}p{0.58\linewidth}@{}}
    \toprule
    Setting & Value \\
    \midrule
    Transformer layers & 1 \\
    Attention heads & 4 \\
    Model dimension & 128 \\
    MLP width & 512 \\
    Activation & ReLU \\
    Layer normalization & None \\
    \bottomrule
  \end{tabular}
\end{table}

\begin{table}[htbp]
  \centering
  \setlength{\belowcaptionskip}{6pt}
  \caption{\textbf{Training configuration for modular addition.}}
  \label{tab:grokking-training-setup}
  \small
  \renewcommand{\arraystretch}{1.12}
  \begin{tabular}{@{}p{0.36\linewidth}p{0.58\linewidth}@{}}
    \toprule
    Setting & Value \\
    \midrule
    Task & $y=(a+b)\bmod 113$ \\
    Train/test pairs & 3,830 / 8,939 (30\% / 70\%) \\
    Optimizer & Full-batch AdamW \\
    Learning rate & $10^{-3}$ \\
    Weight decay & 1 \\
    Adam coefficients & $(0.9,0.98)$ \\
    Learning-rate warmup & 10 steps \\
    Training seed & 0 \\
    Representative checkpoint epochs & 1,000; 3,800; 10,000; 25,300 \\
    \bottomrule
  \end{tabular}
\end{table}

\begin{table}[htbp]
  \centering
  \setlength{\belowcaptionskip}{6pt}
  \caption{\textbf{Unlearning configuration for modular addition.} Learning rates are those used for the reported chance-threshold results.}
  \label{tab:grokking-unlearning-setup}
  \small
  \renewcommand{\arraystretch}{1.12}
  \begin{tabular}{@{}p{0.36\linewidth}p{0.58\linewidth}@{}}
    \toprule
    Setting & Value \\
    \midrule
    Optimizer & Full-batch AdamW \\
    Learning rate & $10^{-4}$ (RL); $10^{-4}$ or $3\times10^{-4}$ (GD) \\
    Weight decay & 0 (RL); 1 (GD) \\
    Retain weight $\lambda$ & 0 (RL); 10 (GD) \\
    Evaluation interval (updates) & 1 \\
    Forget/retain examples & 383 / 3,447 \\
    Forget-set selection seeds & 1--5 \\
    Unlearning seed & 0 \\
    Comparison criterion & Accuracy $\leq\tau$ \\
    \bottomrule
  \end{tabular}
\end{table}

Each unlearning run starts from the selected training checkpoint with a fresh optimizer.
The same forget/retain partition is used for both methods and all four checkpoints within each selection seed; the training run, train/test split, and unlearning seed remain fixed.
All 40 (checkpoint, method, seed) combinations in Table~\ref{tab:grokking-chance} reach forget accuracy $\leq\tau$.
We report mean retain accuracy with 95\% Student-$t$ confidence intervals across the five forget-set selection seeds.
The ten-checkpoint trajectory sweep uses a 10\% forget-accuracy threshold; the chance-threshold comparison uses the four representative checkpoints.
The trajectory figure uses forget-set selection seed 1.

\FloatBarrier
\subsection{Circuit Measurements}
\label{app:circuit-measurements}
We compute FVE and restricted loss using the fixed key frequencies $K=\{14,35,42,52,55\}$ identified in the fully trained model.
Let $Z_{ab,c}$ denote the logit for answer $c$ on input $(a,b)$, with $a,b,c\in\{0,\ldots,p-1\}$ and $p=113$.
For each output class, we obtain the restricted logits $Z^K$ by orthogonally projecting the logits onto
\begin{equation}
  \mathcal{S}_K=\operatorname{span}\!\left\{1,\cos\,\!\bigl(\omega_k(a+b)\bigr),\sin\,\!\bigl(\omega_k(a+b)\bigr):k\in K\right\},
  \qquad \omega_k=\frac{2\pi k}{p}.
\end{equation}
The constant component preserves each output class's mean logit $\bar Z_c=p^{-2}\sum_{a,b}Z_{ab,c}$.

\paragraph{FVE.}
We measure the fraction of logit variance explained by the key frequency components as
\begin{equation}
  \mathrm{FVE}
  =\frac{\sum_{a,b,c}(Z^K_{ab,c}-\bar Z_c)^2}
  {\sum_{a,b,c}(Z_{ab,c}-\bar Z_c)^2}.
\end{equation}
The sums range over all input pairs and answer classes.

\paragraph{Restricted loss.}
We compute test cross-entropy using the restricted logits \citep{nanda2023progress}:
\begin{equation}
  \mathcal{L}_{\mathrm{restricted}}
  =-\frac{1}{|\mathcal{D}_{\mathrm{test}}|}
  \sum_{(a,b)\in\mathcal{D}_{\mathrm{test}}}
  \log\frac{\exp Z^K_{ab,y_{ab}}}{\sum_{c=0}^{p-1}\exp Z^K_{ab,c}},
  \qquad y_{ab}=(a+b)\bmod p.
\end{equation}
Here $\mathcal{D}_{\mathrm{test}}$ denotes the test input pairs.
Lower restricted loss indicates that the Fourier circuit can solve the task more effectively on its own.

\FloatBarrier
\subsection{Fourier Gradient Filtering}
\label{app:fourier-projection}
\paragraph{Implementation.}
During RL, we adjust the logit gradient to remove components along the identified Fourier directions before backpropagation.
Let $Z$ denote the output logits.
For each key frequency $k\in K=\{14,35,42,52,55\}$, define a unit-norm direction $E_k[a,b,c]\propto\cos(2\pi k(a+b-c)/p)$ for $c<p$, with a zero column for the extra token.
These five directions are mutually orthogonal.
Given the RL loss gradient $g=\partial\mathcal{L}_{\mathrm{RL}}/\partial Z$, we project it onto the orthogonal complement of their span and backpropagate
\begin{equation}
  g_{\perp}=g-\sum_{k\in K}\langle g,E_k\rangle_F E_k.
\end{equation}
Here $\langle\cdot,\cdot\rangle_F$ denotes the Frobenius inner product.
The key frequency components of the logits represent the circuit's contribution.
Removing the RL gradient along these directions eliminates the direct signal for changing that contribution, reducing pressure to modify the circuit during unlearning.
This, however, does not guarantee complete protection against circuit damage, since the constraint on logit gradients need not be preserved by parameter updates.

\paragraph{Experimental settings.}
We compare unmodified RL, key frequency filtering, and non-key frequency filtering.
All three conditions preserve the original RL objective and its 114-dimensional output vocabulary, with identical incorrect labels and fresh optimizer states within each seed.
The non-key control filters gradient components at five frequencies sampled once from the complement of $K$, using seed 17: $\{20,21,25,28,36\}$.
Training checkpoints, data memberships, and the five $\mathcal{D}_f$ selection seeds match Section~\ref{sec:strategy-damage}.
All conditions use full-batch AdamW with learning rate $10^{-4}$ and zero weight decay, stopping when forget accuracy first reaches $\tau$ or below.
All 60 (checkpoint, seed, condition) combinations reach forget accuracy $\leq\tau$.

\paragraph{Additional result.}
Key frequency filtering substantially improves retain accuracy in the cleanup and grokked checkpoints, where predictions rely strongly on the Fourier circuit, while having much smaller effects during memorization and circuit formation (Table~\ref{tab:projection-phases}).
In the cleanup and grokked checkpoints, these gains accompany higher FVE and lower restricted loss, whereas non-key filtering remains close to unmodified RL.
This dependence on training phase indicates that filtering is most effective when the protected circuit already plays a substantial role in the model's predictions.

\begin{table}[htbp]
  \centering
  \setlength{\belowcaptionskip}{6pt}
  \caption{\textbf{Fourier gradient filtering across training phases.}
  Results at forget accuracy $\leq\tau$, reported as mean $\pm$ 95\% CI across five $\mathcal{D}_f$ selection seeds.}
  \label{tab:projection-phases}
  \small
  \setlength{\tabcolsep}{5pt}
  \renewcommand{\arraystretch}{1.12}
  \begin{tabular}{@{}lccc@{}}
    \toprule
    Intervention & Retain accuracy (\%) & FVE & Restricted loss \\
    \midrule
    \multicolumn{4}{@{}l}{\textbf{Memorization}} \\
    RL & 59.5\,\textpm{}\,3.3 & 0.0104\,\textpm{}\,0.0005 & 5.382\,\textpm{}\,0.448 \\
    RL + filtering (non-key) & 59.7\,\textpm{}\,3.1 & 0.0104\,\textpm{}\,0.0005 & 5.384\,\textpm{}\,0.465 \\
    RL + filtering (key) & 60.2\,\textpm{}\,2.6 & 0.0124\,\textpm{}\,0.0004 & 5.296\,\textpm{}\,0.473 \\
    \midrule
    \multicolumn{4}{@{}l}{\textbf{Circuit formation}} \\
    RL & 52.4\,\textpm{}\,7.3 & 0.0239\,\textpm{}\,0.0029 & 4.653\,\textpm{}\,0.142 \\
    RL + filtering (non-key) & 52.1\,\textpm{}\,6.2 & 0.0237\,\textpm{}\,0.0025 & 4.664\,\textpm{}\,0.156 \\
    RL + filtering (key) & 55.5\,\textpm{}\,4.7 & 0.0312\,\textpm{}\,0.0024 & 4.404\,\textpm{}\,0.126 \\
    \midrule
    \multicolumn{4}{@{}l}{\textbf{Cleanup}} \\
    RL & 14.2\,\textpm{}\,1.6 & 0.1064\,\textpm{}\,0.0185 & 1.484\,\textpm{}\,0.237 \\
    RL + filtering (non-key) & 14.5\,\textpm{}\,1.4 & 0.1094\,\textpm{}\,0.0181 & 1.442\,\textpm{}\,0.221 \\
    RL + filtering (key) & 51.8\,\textpm{}\,2.4 & 0.4533\,\textpm{}\,0.0064 & 0.038\,\textpm{}\,0.026 \\
    \midrule
    \multicolumn{4}{@{}l}{\textbf{Grokked}} \\
    RL & 8.1\,\textpm{}\,0.7 & 0.1861\,\textpm{}\,0.0258 & 2.324\,\textpm{}\,0.183 \\
    RL + filtering (non-key) & 8.2\,\textpm{}\,0.7 & 0.1873\,\textpm{}\,0.0271 & 2.316\,\textpm{}\,0.187 \\
    RL + filtering (key) & 30.6\,\textpm{}\,4.1 & 0.5085\,\textpm{}\,0.0204 & 0.167\,\textpm{}\,0.167 \\
    \bottomrule
  \end{tabular}
\end{table}

\subsection{Causal Intervention}
\label{app:causal-intervention}
We intervene on the RL logit gradient: before backpropagation, we project it onto the orthogonal complement of the subspace spanned by the five global directions $\cos(\omega_k(a+b-c))$ associated with the identified circuit.
This filtering removes the gradient signal for changing the Fourier circuit's contribution to the logits, encouraging its preservation during unlearning.
We compare this intervention with unmodified RL and filtering of the same number of non-key frequency directions, using the same four models, five $\mathcal{D}_f$ selection seeds, and forget-accuracy threshold $\tau$ as in Section~\ref{sec:strategy-damage}.

In the grokked model, this intervention yielded higher FVE (0.509 vs. 0.186) and lower restricted loss (0.167 vs. 2.324; Table~\ref{tab:circuit-projection}), indicating better preservation of the Fourier circuit.
Accordingly, mean retain accuracy increased from 8.1\% to 30.6\%, indicating that disruption of this shared solution mechanism contributes to retain damage during unlearning.
The non-key control left FVE and restricted loss nearly unchanged, showing no comparable circuit protection, and retain accuracy correspondingly remained at 8.2\%.
In summary, these findings show why selective forgetting becomes difficult when the targeted and retained examples rely on the same solution mechanism.
See Appendices~\ref{app:circuit-measurements} and~\ref{app:fourier-projection} for details of the circuit measurements and the intervention, respectively.


\begin{table}[!htbp]
  \caption{\textbf{Protecting the Fourier circuit reduces retain damage.}
  Unlearning (RL) results for the grokked model at forget accuracy $\leq\tau$,
  using the same optimizer settings across conditions.
  Filtering removes logit-gradient components along the specified Fourier directions.
  Entries are mean $\pm$ 95\% CI across five $\mathcal{D}_f$ selection seeds.}
  \label{tab:circuit-projection}
  \begin{center}
  \small
  \setlength{\tabcolsep}{6pt}
  \begin{tabular}{@{}lccc@{}}
    \toprule
    Intervention & Retain acc. (\%) & FVE & Restricted loss \\
    \midrule
    RL & 8.1\,\textpm{}\,0.7 & 0.186\,\textpm{}\,0.026 & 2.324\,\textpm{}\,0.183 \\
    RL + filtering (non-key) & 8.2\,\textpm{}\,0.7 & 0.187\,\textpm{}\,0.027 & 2.316\,\textpm{}\,0.187 \\
    RL + filtering (key) & 30.6\,\textpm{}\,4.1 & 0.509\,\textpm{}\,0.020 & 0.167\,\textpm{}\,0.167 \\
    \bottomrule
  \end{tabular}
  \end{center}
\end{table}

\FloatBarrier
\section{Bucketed Modular Addition Experiments}
\label{app:bma}
\subsection{Configuration}
\label{app:bma-config}
The model architecture is the one used for modular addition (Table~\ref{tab:grokking-architecture}), with the model dimension and MLP width doubled to 256 and 1,024.
Tables~\ref{tab:bma-training-setup} and~\ref{tab:bma-unlearning-setup} summarize the training and unlearning configurations, respectively.
The five bucket sizes share the same ordered input pairs and train/test split, and each draws its within-bucket assignments once before training.
Each unlearning run starts from the epoch-30,000 checkpoint with a fresh optimizer.
All five checkpoints reach 100\% training accuracy.
Test bucket accuracy is 93.57\%, 93.82\%, and 96.63\% for $M=4,16,64$, while test within-bucket accuracy---the fraction of test examples whose highest-probability answer within the correct bucket carries the correct index---is 25.09\%, 5.98\%, and 1.63\%, close to the $1/M$ chance levels of 25\%, 6.25\%, and 1.56\%.
The models therefore generalize the modular rule but not the index, which is what makes $\log_2 M$ the memorization floor.

\begin{table}[htbp]
  \centering
  \setlength{\belowcaptionskip}{6pt}
  \caption{\textbf{Training configuration for bucketed modular addition.} Settings are fixed across bucket sizes.}
  \label{tab:bma-training-setup}
  \small
  \renewcommand{\arraystretch}{1.12}
  \begin{tabular}{@{}p{0.42\linewidth}p{0.52\linewidth}@{}}
    \toprule
    Setting & Value \\
    \midrule
    Task & $y=M((a+b)\bmod p)+u_{a,b}$, $pM=256$ \\
    Bucket size $M$ / Memorization Floor $\log_2 M$ & $1,4,16,64,256$ / $0,2,4,6,8$ bits \\
    Output classes & 256 \\
    Train/test pairs & 19,660 / 45,876 (30\% / 70\%) \\
    Optimizer & Full-batch AdamW \\
    Learning rate & $10^{-3}$ \\
    Weight decay & 1 \\
    Adam coefficients & $(0.9,0.98)$ \\
    Learning-rate warmup & 10 steps \\
    Unlearning checkpoint epoch & 30,000 \\
    \bottomrule
  \end{tabular}
\end{table}


\begin{table}[htbp]
  \centering
  \setlength{\belowcaptionskip}{6pt}
  \caption{\textbf{Unlearning configuration for bucketed modular addition.} Settings are fixed across bucket sizes and selection seeds; only GD depends on the forget fraction.}
  \label{tab:bma-unlearning-setup}
  \small
  \renewcommand{\arraystretch}{1.12}
  \begin{tabular}{@{}p{0.36\linewidth}p{0.58\linewidth}@{}}
    \toprule
    Setting & Value \\
    \midrule
    Optimizer / weight decay & Full-batch AdamW / 0 \\
    Forget/retain examples & 197 / 19,463 (1\%); 983 / 18,677 (5\%); 1,966 / 17,694 (10\%) \\
    Forget-set selection seeds & 17, 26, 42 \\
    Stopping rule & $P_f\leq0.0039$, or update budget \\
    Comparison criterion & $P_f=0.1$ \\
    \bottomrule
  \end{tabular}

  \vspace{5pt}
  \begin{tabular*}{0.97\linewidth}{@{\extracolsep{\fill}}llccc@{}}
    \toprule
    Method & Forget fraction & Learning rate & Retain weight $\lambda$ & $\beta$ \\
    \midrule
    GD & 1\% & $5\times10^{-5}$ & 10 & --- \\
    GD & 5\%, 10\% & $10^{-4}$ & 10 & --- \\
    RL & All & $2\times10^{-5}$ & 0 & --- \\
    DPO & All & $5\times10^{-6}$ & 0 & 0.1 \\
    ME & All & $10^{-4}$ & 0 & --- \\
    \bottomrule
  \end{tabular*}
\end{table}

We unlearn with GD, RL, DPO, and ME as defined in Appendix~\ref{app:unlearning-methods}.
Within each forget fraction and seed, the same input pairs form the forget and retain sets across methods and bucket sizes, and the seed also fixes the incorrect labels that RL and DPO train toward.
Evaluations occur every two updates, or every four for GD at 5\% and 10\%.
A run stops at mean forget-answer probability $\leq0.0039$ or an update budget, which is separate from the comparison at $P_f=0.1$.
For each run we take the first pair of adjacent saved evaluations bracketing $P_f=0.1$, linearly interpolate $P_r$ as a function of $P_f$ between them, and report retain damage $1-P_r$; all 180 method--bucket--fraction--seed combinations cross this threshold.
Figure~\ref{fig:bma-excess} averages this damage over the three seeds with 95\% confidence intervals, which describe seed variation for fixed trained models and are not truncated to $[0,1]$.
Initial $P_r$ exceeds $0.999997$, so $1-P_r$ closely approximates the decrease from the pre-unlearning model.

\subsection{Memorization Floor and Retain Damage Results}
\label{app:bma-damage-results}
\begin{figure}[!htbt]
  \centering
  \includegraphics[width=\linewidth]{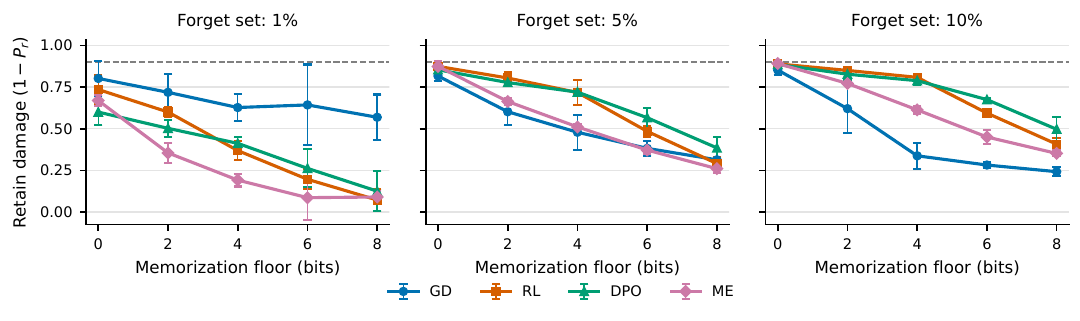}
  \caption{\textbf{A lower memorization floor leaves greater retain damage.}
  Retain damage $1-P_r$ at $P_f=0.1$. Points and error bars show means and 95\% CIs across
  three $\mathcal{D}_f$ selection seeds. Dashed lines mark damage $0.9$, the
  nonselective baseline where $P_f=P_r=0.1$.}
  \label{fig:bma-excess}
\end{figure}

Figure~\ref{fig:bma-excess} reports retain damage at $P_f=0.1$ for each method, bucket size, and forget fraction.
Damage falls as memorization floor grows in all three panels.
At zero bits, where the rule determines the whole answer, every method sits just below the dashed nonselective baseline at $0.9$, the damage that follows when unlearning drives $P_r$ down to $P_f$; by eight bits all four have moved well clear of it.
A larger forget fraction leaves more damage at a given bucket size, so the three panels sit at successively higher levels.
Appendix~\ref{app:bma-cor} quantifies the association.

\subsection{Correlation Analysis of Memorization Floor and Retain Damage}
\label{app:bma-cor}
Table~\ref{tab:bma-correlations} reports Spearman's $\rho_s$ and Pearson's $r$ between memorization floor ($\log_2 M$, in bits) and the five unrounded mean damages, computed separately within each method and forget fraction.
With $n=5$ bucket sizes per cell, they describe monotonic and linear association across bucket-size means, not variability across seeds.
Ten of the twelve cells reach $\rho_s=-1.000$: at 5\% and 10\%, mean damage decreases with every increase in bucket size for all four methods.
The exceptions are GD and ME at 1\%, both $\rho_s=-0.900$, and each comes from a single reversal---GD from $0.627$ at $M=16$ to $0.643$ at $M=64$, and ME from $0.086$ at $M=64$ to $0.090$ at $M=256$---while RL and DPO decrease monotonically there as well.
Pearson's $r$ ranges from $-0.924$ to $-0.997$, so damage falls close to linearly in bits rather than in $M$.
The sweep varies the granularity of the shared rule together with the required memorization; it supports the predicted association in this task family rather than a method-independent bound on retain damage.

\begin{table}[!htbp]
  \setlength{\belowcaptionskip}{6pt}
  \centering
  \caption{\textbf{Memorization floor and retain damage.} Spearman's $\rho_s$
  and Pearson's $r$ between memorization floor ($\log_2 M$) and mean retain
  damage ($1-P_r$) at $P_f=0.1$, across five bucket sizes ($n=5$),
  calculated separately
  for each forget fraction and method.}
  \label{tab:bma-correlations}
  \small
  \setlength{\tabcolsep}{6pt}
  \begin{tabular}{@{}clrr@{}}
    \toprule
    Forget & Method & $\rho_s$ & $r$ \\
    \midrule
    1\% & GD & $-$0.900 & $-$0.950 \\
     & RL & $-$1.000 & $-$0.995 \\
     & DPO & $-$1.000 & $-$0.994 \\
     & ME & $-$0.900 & $-$0.924 \\
    \midrule
    5\% & GD & $-$1.000 & $-$0.975 \\
     & RL & $-$1.000 & $-$0.972 \\
     & DPO & $-$1.000 & $-$0.973 \\
     & ME & $-$1.000 & $-$0.993 \\
    \midrule
    10\% & GD & $-$1.000 & $-$0.942 \\
     & RL & $-$1.000 & $-$0.946 \\
     & DPO & $-$1.000 & $-$0.957 \\
     & ME & $-$1.000 & $-$0.997 \\
    \bottomrule
  \end{tabular}
\end{table}

\FloatBarrier
\section{LLM Experiments}
\label{app:llm}
This section covers both LLM settings: verbatim recall of inserted passages (Appendix~\ref{app:verbatim-recall-unlearning}) and factual recall of inserted biographies (Appendix~\ref{app:factual-recall-unlearning}).
They share the Hubble models described next, and differ in corpus, scoring, and forget-set construction.

\subsection{Models and Data}
\label{app:hubble}
Hubble releases each model as a pair pretrained on the same base corpus: the \emph{perturbed} model additionally sees the controlled insertions, and the \emph{standard} model does not.
We unlearn from the perturbed model and take the standard model as the reference likelihood, since the two differ only in those insertions.
One standard model serves every experiment at a given size and budget; it is not retrained per forget set.
Table~\ref{tab:hubble-models} lists the four pairs and Table~\ref{tab:hubble-data} the inserted we use, of which Gutenberg Popular and Wikipedia supply the passages of Appendix~\ref{app:verbatim-recall-unlearning} and the YAGO biographies the facts of Appendix~\ref{app:factual-recall-unlearning}.
The frozen pre-unlearning model used by preference objectives is the initial perturbed checkpoint, not the standard model.

\begin{table}[htbp]
  \centering
  \setlength{\belowcaptionskip}{6pt}
  \caption{\textbf{Hubble model pairs.} The two members of a pair share a pretraining procedure and differ only in whether the controlled insertions are present.}
  \label{tab:hubble-models}
  \small
  \renewcommand{\arraystretch}{1.12}
  \begin{tabular}{@{}lll@{}}
    \toprule
    Size & Pretraining budget & Repository \\
    \midrule
    8B & 100B tokens & \texttt{allegrolab/hubble-8b-100b\_toks-\{%
      \href{https://huggingface.co/allegrolab/hubble-8b-100b_toks-standard-hf}{standard},%
      \href{https://huggingface.co/allegrolab/hubble-8b-100b_toks-perturbed-hf}{perturbed}\}-hf} \\
    8B & 500B tokens & \texttt{allegrolab/hubble-8b-500b\_toks-\{%
      \href{https://huggingface.co/allegrolab/hubble-8b-500b_toks-standard-hf}{standard},%
      \href{https://huggingface.co/allegrolab/hubble-8b-500b_toks-perturbed-hf}{perturbed}\}-hf} \\
    1B & 100B tokens & \texttt{allegrolab/hubble-1b-100b\_toks-\{%
      \href{https://huggingface.co/allegrolab/hubble-1b-100b_toks-standard-hf}{standard},%
      \href{https://huggingface.co/allegrolab/hubble-1b-100b_toks-perturbed-hf}{perturbed}\}-hf} \\
    1B & 500B tokens & \texttt{allegrolab/hubble-1b-500b\_toks-\{%
      \href{https://huggingface.co/allegrolab/hubble-1b-500b_toks-standard-hf}{standard},%
      \href{https://huggingface.co/allegrolab/hubble-1b-500b_toks-perturbed-hf}{perturbed}\}-hf} \\
    \bottomrule
  \end{tabular}
\end{table}

\begin{table}[htbp]
  \centering
  \setlength{\belowcaptionskip}{6pt}
  \caption{\textbf{Inserted corpora.} The Hubble insertions these experiments use. Each is added to the base corpus when the perturbed model is pretrained and is absent from the standard model; Hubble inserts further corpora that we do not use.}
  \label{tab:hubble-data}
  \small
  \renewcommand{\arraystretch}{1.12}
  \begin{tabular}{@{}l@{\hspace{38pt}}l@{}}
    \toprule
    Corpus & Dataset \\
    \midrule
    Gutenberg Popular & \href{https://huggingface.co/datasets/allegrolab/passages_gutenberg_popular}{\texttt{allegrolab/passages\_gutenberg\_popular}} \\
    Wikipedia & \href{https://huggingface.co/datasets/allegrolab/passages_wikipedia}{\texttt{allegrolab/passages\_wikipedia}} \\
    YAGO biographies & \href{https://huggingface.co/datasets/allegrolab/biographies_yago}{\texttt{allegrolab/biographies\_yago}} \\
    \bottomrule
  \end{tabular}
\end{table}

\subsection{Verbatim Recall}
\label{app:verbatim-recall-unlearning}
\subsubsection{Configuration}
\label{app:verbatim-config}
Passages are scored under teacher forcing, with Eq.~\ref{eq:verbatim-prob} taken over every token of the passage, each conditioned on the \texttt{<BOS>} separator and the preceding tokens.
Ranking the 80 passages of the $256\times$ Gutenberg Popular tier by estimated memorization floor and taking the top and bottom 30\% gives the High and Low forget sets, leaving the middle 40\% as the ID-seen retain group; Table~\ref{tab:verbatim-groups} lists the resulting groups with the passage counts each model contributes.
Memorization floor is summed over a passage's tokens, so a longer passage accumulates more of it simply by being longer; \citet{morris2026memorize} sidestep this by scoring fixed-length sequences, whereas our passages vary in length, which correlates with memorization floor at Spearman $0.55$.
Greedy swaps between the two tails therefore equalize their token totals, which differ by 7 tokens out of 6{,}259 for Hubble-8B-100B, so that the split separates passages by the memorization floor they carry rather than by how much text they contain.
Table~\ref{tab:verbatim-floors} reports what the split achieves: the High set carries a higher memorization floor than the Low set in every model, by roughly 250 bits per passage, and the two intervals never overlap.
For Hubble-1B-500B the tier is first restricted to passages with $\bar{P}_{\mathrm{pert}}\geq0.6$: there the 256-times insertions did not saturate, and without this restriction the memorization floor ranking would also rank passages by how much was memorized at all.

\begin{table}[htbp]
  \centering
  \setlength{\belowcaptionskip}{6pt}
  \caption{\textbf{Forget and retain groups for verbatim recall.} Entries are passage counts. High and Low are the top and bottom 30\% of the $256\times$ Gutenberg Popular tier by memorization floor. For Hubble-1B-500B the tier is first restricted to $\bar{P}_{\mathrm{pert}}\geq0.6$, which shrinks every seen group.}
  \label{tab:verbatim-groups}
  \small
  \setlength{\tabcolsep}{4pt}
  \renewcommand{\arraystretch}{1.12}
  \begin{tabular}{@{}llrrrrl@{}}
    \toprule
    Role & Group & \shortstack{Hubble\\8B-100B} & \shortstack{Hubble\\8B-500B} & \shortstack{Hubble\\1B-100B} & \shortstack{Hubble\\1B-500B} & Source and selection \\
    \midrule
    Forget & High & 24 & 24 & 24 & 18 & Gutenberg Popular, top 30\% \\
    Forget & Low & 24 & 24 & 24 & 18 & Gutenberg Popular, bottom 30\% \\
    \midrule
    Retain & ID-seen & 32 & 32 & 32 & 21 & Gutenberg Popular, middle 40\% \\
    Retain & OOD-seen & 76 & 76 & 76 & 16 & Wikipedia, $256\times$ inserted \\
    Retain & Unseen & 400 & 400 & 400 & 400 & 200 held-out passages of each corpus \\
    \bottomrule
  \end{tabular}
\end{table}

\begin{table}[htbp]
  \centering
  \setlength{\belowcaptionskip}{6pt}
  \caption{\textbf{Memorization floor of the verbatim forget sets.} Per-passage floor in bits, as mean $\pm$ 95\% CI over the passages of the group.}
  \label{tab:verbatim-floors}
  \small
  \setlength{\tabcolsep}{4pt}
  \renewcommand{\arraystretch}{1.12}
  \begin{tabular}{@{}lrrrr@{}}
    \toprule
    Group & \shortstack{Hubble\\8B-100B} & \shortstack{Hubble\\8B-500B} & \shortstack{Hubble\\1B-100B} & \shortstack{Hubble\\1B-500B} \\
    \midrule
    High & 1136\,\textpm{}\,89 & 1048\,\textpm{}\,88 & 1233\,\textpm{}\,96 & 1062\,\textpm{}\,97 \\
    Low & 881\,\textpm{}\,82 & 780\,\textpm{}\,88 & 985\,\textpm{}\,83 & 838\,\textpm{}\,80 \\
    \bottomrule
  \end{tabular}
\end{table}

\begin{table}[htbp]
  \centering
  \setlength{\belowcaptionskip}{6pt}
  \caption{\textbf{Unlearning hyperparameters for verbatim recall.} Each run uses AdamW, gradient clipping at 1.0, and one pass over its forget set per epoch. NPO uses $\beta=0.1$.}
  \label{tab:verbatim-unlearning-setup}
  \small
  \renewcommand{\arraystretch}{1.12}
  \begin{tabular}{@{}lcl@{}}
    \toprule
    Method & Learning rate & Retain objective \\
    \midrule
    GA & $10^{-5}$ & None \\
    GD & $10^{-5}$ & Cross-entropy on ID-seen ($\lambda=1$) \\
    NPO & $10^{-5}$ & None \\
    CEU & $10^{-5}$ & None \\
    \bottomrule
  \end{tabular}
\end{table}

Each forget set is unlearned from a fresh copy of the perturbed checkpoint with the settings in Table~\ref{tab:verbatim-unlearning-setup}.
GD anchors its retain term on the ID-seen group itself, so its ID-seen entries are a cheating baseline while its OOD-seen and unseen entries remain independent measurements.
Every retain group is evaluated after each epoch, and a run stops at the first epoch with $\bar{P}_f\leq0.1$, so runs stop at different values below the threshold.
We therefore compare $\bar{P}_r$ at exactly $\bar{P}_f=0.1$, linearly interpolating between the stopping epoch and the one before it.
Reported intervals are 95\% confidence intervals over the retain passages of a group, which both conditions share; they describe passage-level variation for a fixed split and model, not variation across forget-set selections or training runs.

\FloatBarrier
\begin{figure}[!htbp]
  \centering
  \setlength{\abovecaptionskip}{6pt}
  \includegraphics[width=\linewidth]{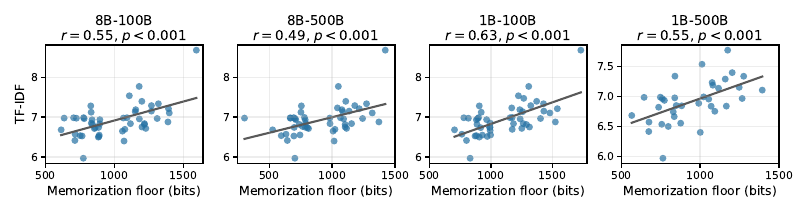}
  \caption{\textbf{Document-level TF-IDF and memorization floor.}
  Each point is a passage of the two forget sets; panel titles give the Pearson
  correlation and its $p$-value.}
  \label{fig:verbatim-tfidf}
\end{figure}
\begin{table}[!htbp]
  \centering
  \setlength{\belowcaptionskip}{6pt}
  \caption{\textbf{Retain damage in verbatim recall.} Hubble-8B and Hubble-1B, $256\times$ Gutenberg Popular. Damage is $100(\bar{P}_{r,\mathrm{before}}-\bar{P}_{r,\mathrm{after}})$ at $\bar{P}_f=0.1$, as mean $\pm$ 95\% CI over the retain passages of a group. Bold marks the smaller damage of a High--Low pair.}
  \label{tab:verbatim-results}
  \scriptsize
  \setlength{\tabcolsep}{2.5pt}
  \renewcommand{\arraystretch}{1.3}
  \begin{tabular*}{\textwidth}{@{\extracolsep{\fill}}clrrrrrrrr@{}}
    \toprule\noalign{\vskip -1pt}
     & & \multicolumn{8}{c}{Retain damage (\%) $\downarrow$} \\[-1.5pt]
    \cmidrule(l){3-10}\noalign{\vskip -1pt}
    \multirow{2}{*}{Model} & \multirow{2}{*}{Retain} & \multicolumn{2}{c}{GA} & \multicolumn{2}{c}{GD} & \multicolumn{2}{c}{NPO} & \multicolumn{2}{c}{CEU} \\[-1.5pt]
    \cmidrule(lr){3-4} \cmidrule(lr){5-6} \cmidrule(lr){7-8} \cmidrule(lr){9-10}\noalign{\vskip -1pt}
    & & \multicolumn{1}{c}{High} & \multicolumn{1}{c}{Low} & \multicolumn{1}{c}{High} & \multicolumn{1}{c}{Low} & \multicolumn{1}{c}{High} & \multicolumn{1}{c}{Low} & \multicolumn{1}{c}{High} & \multicolumn{1}{c}{Low} \\[-1.5pt]
    \midrule
    \multirow{3}{*}{\rotatebox[origin=c]{90}{\shortstack{\textbf{Hubble}\\\textbf{8B-100B}}}} & ID-seen & \textbf{45.5\,\textpm{}\,3.8} & 54.0\,\textpm{}\,3.2 & \textbf{17.3\,\textpm{}\,2.3} & 17.9\,\textpm{}\,1.8 & \textbf{45.0\,\textpm{}\,3.8} & 52.8\,\textpm{}\,3.3 & \textbf{54.2\,\textpm{}\,6.5} & 74.0\,\textpm{}\,5.1 \\
     & OOD-seen & \textbf{3.4\,\textpm{}\,0.5} & 5.4\,\textpm{}\,0.9 & \textbf{2.9\,\textpm{}\,0.5} & 4.0\,\textpm{}\,0.7 & \textbf{3.4\,\textpm{}\,0.5} & 5.2\,\textpm{}\,0.8 & \textbf{9.8\,\textpm{}\,3.0} & 18.7\,\textpm{}\,4.3 \\
     & Unseen & \textbf{0.5\,\textpm{}\,0.1} & 0.7\,\textpm{}\,0.1 & \textbf{0.6\,\textpm{}\,0.1} & 0.8\,\textpm{}\,0.1 & \textbf{0.5\,\textpm{}\,0.1} & 0.7\,\textpm{}\,0.1 & $-$0.6\,\textpm{}\,0.1 & $-$0.4\,\textpm{}\,0.1 \\
    \midrule
    \multirow{3}{*}{\rotatebox[origin=c]{90}{\shortstack{\textbf{Hubble}\\\textbf{8B-500B}}}} & ID-seen & \textbf{14.8\,\textpm{}\,2.1} & 26.4\,\textpm{}\,2.9 & \textbf{0.2\,\textpm{}\,0.6} & 3.6\,\textpm{}\,1.1 & \textbf{13.4\,\textpm{}\,2.1} & 25.2\,\textpm{}\,2.9 & \textbf{44.1\,\textpm{}\,6.1} & 63.1\,\textpm{}\,4.2 \\
     & OOD-seen & \textbf{2.5\,\textpm{}\,0.5} & 2.8\,\textpm{}\,0.6 & \textbf{2.0\,\textpm{}\,0.4} & 2.2\,\textpm{}\,0.6 & \textbf{2.3\,\textpm{}\,0.5} & 2.7\,\textpm{}\,0.6 & \textbf{4.7\,\textpm{}\,1.1} & 5.1\,\textpm{}\,1.1 \\
     & Unseen & \textbf{0.2\,\textpm{}\,0.1} & 0.3\,\textpm{}\,0.1 & 0.3\,\textpm{}\,0.1 & 0.3\,\textpm{}\,0.1 & \textbf{0.2\,\textpm{}\,0.1} & 0.3\,\textpm{}\,0.1 & $-$0.1\,\textpm{}\,0.1 & 0.3\,\textpm{}\,0.1 \\
    \midrule
    \multirow{3}{*}{\rotatebox[origin=c]{90}{\shortstack{\textbf{Hubble}\\\textbf{1B-100B}}}} & ID-seen & \textbf{48.0\,\textpm{}\,3.9} & 55.9\,\textpm{}\,3.6 & \textbf{16.9\,\textpm{}\,1.8} & 20.1\,\textpm{}\,1.9 & \textbf{47.7\,\textpm{}\,3.8} & 55.7\,\textpm{}\,3.6 & 59.7\,\textpm{}\,4.8 & \textbf{58.4\,\textpm{}\,3.3} \\
     & OOD-seen & \textbf{6.4\,\textpm{}\,0.9} & 11.2\,\textpm{}\,1.4 & \textbf{5.3\,\textpm{}\,0.7} & 9.5\,\textpm{}\,1.1 & \textbf{6.3\,\textpm{}\,0.9} & 11.2\,\textpm{}\,1.4 & 18.0\,\textpm{}\,3.7 & \textbf{9.5\,\textpm{}\,1.2} \\
     & Unseen & $-$0.2\,\textpm{}\,0.0 & 0.2\,\textpm{}\,0.0 & $-$0.1\,\textpm{}\,0.0 & 0.8\,\textpm{}\,0.1 & $-$0.1\,\textpm{}\,0.0 & 0.2\,\textpm{}\,0.0 & $-$0.2\,\textpm{}\,0.1 & 0.6\,\textpm{}\,0.1 \\
    \midrule
    \multirow{3}{*}{\rotatebox[origin=c]{90}{\shortstack{\textbf{Hubble}\\\textbf{1B-500B}}}} & ID-seen & \textbf{22.6\,\textpm{}\,2.7} & 25.9\,\textpm{}\,2.4 & $-$2.9\,\textpm{}\,3.1 & $-$3.3\,\textpm{}\,2.8 & \textbf{22.2\,\textpm{}\,2.7} & 25.7\,\textpm{}\,2.4 & \textbf{34.7\,\textpm{}\,3.6} & 45.8\,\textpm{}\,3.1 \\
     & OOD-seen & \textbf{2.7\,\textpm{}\,0.5} & 3.6\,\textpm{}\,0.9 & \textbf{2.3\,\textpm{}\,0.6} & 3.4\,\textpm{}\,0.9 & \textbf{2.6\,\textpm{}\,0.6} & 3.6\,\textpm{}\,0.9 & \textbf{2.3\,\textpm{}\,0.8} & 4.2\,\textpm{}\,1.0 \\
     & Unseen & \textbf{0.2\,\textpm{}\,0.1} & 0.5\,\textpm{}\,0.1 & \textbf{0.2\,\textpm{}\,0.0} & 0.4\,\textpm{}\,0.1 & \textbf{0.2\,\textpm{}\,0.1} & 0.5\,\textpm{}\,0.1 & $-$0.0\,\textpm{}\,0.1 & 0.1\,\textpm{}\,0.1 \\
    \bottomrule
  \end{tabular*}
\end{table}

\subsubsection{Lexical Rarity and Memorization Floor}
\label{app:verbatim-rarity}
We measure lexical rarity with the TF-IDF statistic of \citet{morris2026memorize},
\begin{equation}
  \mathrm{TF\text{-}IDF}(d;D)=\frac{1}{|d|}\sum_{w\in d}\log\frac{|D|}{\mathrm{tf}(w,D)},
  \label{eq:tfidf-main}
\end{equation}
where the sum runs over the word occurrences of passage $d$, $\mathrm{tf}(w,D)$ counts the occurrences of $w$ in the corpus $D$, and $|d|$ and $|D|$ are their word counts.
It departs from the usual document-level score in judging a word's frequency against the corpus rather than the passage; we take $D$ to be the 680 inserted Gutenberg Popular training passages.
Figure~\ref{fig:verbatim-tfidf} plots TF-IDF against estimated memorization floor for all four models, with the Pearson correlation and its $p$-value in each panel title; memorization floor and TF-IDF rise together in every panel.
Every passage plotted was inserted 256 times, so the spread is not a duplication effect.

Table~\ref{tab:verbatim-examples} shows what the statistic picks up on: one passage from each forget set of Hubble-8B-100B, matched in length so that their memorization floor totals are directly comparable.
Highlighting marks the words occurring five or fewer times in the 680 inserted Gutenberg Popular passages.
The high passage is dense with such words---\emph{Humelbergius}, \emph{Z\"urich}, \emph{antiquary}---while the low passage is assembled almost entirely from common vocabulary.

\subsubsection{Memorization Floor and Retain Damage Results}
\label{app:verbatim-damage-results}
Table~\ref{tab:verbatim-results} collects retain damage for all four models.
The Low forget set leaves greater damage than the High one at both model sizes and both pretraining budgets, for nearly every method and retain group, so the direction reported in Section~\ref{sec:verbatim} does not depend on scale.
Damage also orders by how much a retain group shares with the forget set: it is largest on the ID-seen passages, which come from the same tier of the same corpus, smaller on the OOD-seen ones, and at most a fraction of a percent on the unseen passages.
This is the ordering the account of Section~\ref{sec:verbatim} predicts, since a retained passage is exposed only through the regularities it shares with what is being removed.

\subsection{Factual Recall}
\label{app:factual-recall-unlearning}
\subsubsection{Configuration}
\label{app:factual-config}
Each inserted biography gives every attribute its own sentence, and a query is the opening of one such sentence, so answering it means completing text the perturbed model saw during pretraining.
Table~\ref{tab:yago-prompts} gives that opening for each relation, together with its candidate set.
The 15 distractors are values of the same relation taken from non-inserted biographies, and a fact keeps the same candidates throughout unlearning.
A candidate is scored by summing its token log probabilities over the answer span, without averaging by answer length.
The reference-relative gain can exceed the four bits of a uniform candidate index when the standard model assigns the correct answer less than $1/16$ probability.
We keep the per-fact gain signed before averaging; its positive part is the estimator of Section~\ref{sec:background-generalization} applied to a choice among candidates, which we do not equate with an information gain over raw answer strings.
For each model and relation we take 50 facts the perturbed model already recalls correctly and split them with selection seed 42 into 15 forget facts and 35 retain facts, held fixed across methods.

\begin{table}[htbp]
  \centering
  \setlength{\belowcaptionskip}{6pt}
  \caption{\textbf{Relations, prompts, and memorization floors.} Each prompt opens the biography sentence that states the attribute, with the subject's name in place of \texttt{\{subject\}}, and is completed by the attribute value. Floors are mean $\pm$ 95\% CI over the relation's selected facts.}
  \label{tab:yago-prompts}
  \small
  \renewcommand{\arraystretch}{1.12}
  \begin{tabular*}{\textwidth}{@{\extracolsep{\fill}}llcc@{}}
    \toprule
    & & \multicolumn{2}{c}{Memorization floor (bits)} \\
    \cmidrule(l){3-4}
    Relation & Prompt & Hubble-8B-100B & Hubble-8B-500B \\
    \midrule
    Email & \texttt{\{subject\} receives email at} & 0.93\,\textpm{}\,0.64 & 0.43\,\textpm{}\,0.42 \\
    Nationality & \texttt{\{subject\} is from} & 1.61\,\textpm{}\,0.44 & 1.09\,\textpm{}\,0.31 \\
    Birthplace & \texttt{\{subject\} was born in} & 4.30\,\textpm{}\,2.05 & 2.30\,\textpm{}\,1.31 \\
    UUID & \texttt{\{subject\} has the unique identifier} & 5.00\,\textpm{}\,0.62 & 5.70\,\textpm{}\,0.70 \\
    \bottomrule
  \end{tabular*}
\end{table}

\begin{table}[htbp]
  \centering
  \setlength{\belowcaptionskip}{6pt}
  \caption{\textbf{Unlearning settings for factual recall.} Only these three settings depend on model size; both sizes use the 100B and 500B pretraining budgets, 15 forget facts out of 50 with selection seed 42, GD retain weight 5, and $\beta=0.1$ for NPO and DPO. Settings are shared by all methods and relations.}
  \label{tab:factual-unlearning-setup}
  \small
  \renewcommand{\arraystretch}{1.12}
  \begin{tabular}{@{}lcc@{}}
    \toprule
    Setting & Hubble-1B & Hubble-8B \\
    \midrule
    Learning rate & $3\times10^{-7}$ & $10^{-5}$ \\
    Initial update budget & 240 & 480 \\
    Continuation for unmatched runs & None & Up to 20 updates \\
    \bottomrule
  \end{tabular}
\end{table}

Each relation is unlearned from a fresh copy of the perturbed checkpoint with the settings in Table~\ref{tab:factual-unlearning-setup}, with the split fixed across methods within a model and the embeddings and language-model head frozen.
GA and GD average the loss over answer tokens, and NPO and DPO use the summed answer log-likelihood relative to the frozen pre-unlearning model.
We evaluate after every update and report each run at the first update whose forget accuracy is at most 10\%.
Out-relation damage weights the remaining retained relations equally, and both pools include the facts GD uses for its retain term.
The Hubble-8B models are run on all four relations, and the out-relation pool is the other three together with university, which is never itself unlearned; each contributes its 35 retained facts, so the pool holds $4\times35$ facts.
Hubble-1B is run on email, nationality, and UUID only: it recalls too few birthplace facts correctly to assemble a pool of 50, so birthplace is dropped at that scale and the out-relation pool is the other two target relations, a pool of $2\times35$ facts.
Intervals are 95\% confidence intervals over the retained facts, clustered by subject for out-relation damage because one subject contributes facts to several relations.

\subsubsection{Memorization Floor and Retain Damage Results}
\label{app:factual-damage-result}
Table~\ref{tab:yago-results} collects retain damage for all four models.
A relation with a higher memorization floor leaves less in-relation damage: UUID, the highest of the four, is damaged no more than email, the lowest, in 15 of the 16 model--method pairs, so the direction reported in Section~\ref{sec:factual} does not depend on scale.
Damage also stays inside the unlearned relation: averaged over the four methods, $\Delta$ is positive for 13 of the 14 model--relation pairs (Table~\ref{tab:yago-method-mean}), the exception being UUID in Hubble-8B-500B, where every method leaves the retained UUID facts untouched, so even a small out-relation figure turns $\Delta$ negative.
This is the ordering the account of Section~\ref{sec:factual} predicts, since a retained fact is exposed only through the regularities it shares with what is being removed.
The ordering is clearest in the Hubble-8B models and weakest in Hubble-1B-500B, where nationality takes the most damage under every method despite carrying a lower floor than UUID.

\begin{table}[!htbp]
  \centering
  \setlength{\belowcaptionskip}{6pt}
    \caption{\textbf{Retain damage in factual recall.} Hubble-8B and Hubble-1B, $64\times$ YAGO. The memorization floor is in bits and damage is the decrease in accuracy in percentage points, both as mean \textpm{} 95\% CI; in-relation intervals are paired across facts. GA, GD, NPO, and DPO all reach the forget target. Within a model, bold marks the relation on which a method leaves the least in-relation damage.}
  \label{tab:yago-results}
  \scriptsize
  \setlength{\tabcolsep}{2.5pt}
  \renewcommand{\arraystretch}{1.2}
  \begin{tabular*}{\textwidth}{@{\extracolsep{\fill}}clrcccccccc@{}}
    \toprule\noalign{\vskip -1pt}
     & & & \multicolumn{8}{c}{Retain damage (\%p) $\downarrow$} \\[-1.5pt]
    \cmidrule(l){4-11}\noalign{\vskip -1pt}
    \multirow{2}{*}{Model} & \multirow{2}{*}{Relation} & \multicolumn{1}{c}{\multirow{2}{*}{\shortstack[c]{Mem.\\Floor}}} & \multicolumn{2}{c}{GA} & \multicolumn{2}{c}{GD} & \multicolumn{2}{c}{NPO} & \multicolumn{2}{c}{DPO} \\[-1.5pt]
    \cmidrule(lr){4-5}\cmidrule(lr){6-7}\cmidrule(lr){8-9}\cmidrule(lr){10-11}\noalign{\vskip -1pt}
    & & & In-relation & Out-relation & In-relation & Out-relation & In-relation & Out-relation & In-relation & Out-relation \\[-1.5pt]
    \midrule
    \multirow{4}{*}{\rotatebox[origin=c]{90}{\shortstack{\textbf{Hubble}\\\textbf{8B-100B}}}} & Email & 0.93\,\textpm{}\,0.64 & 77.1\,\textpm{}\,14.6 & 3.6\,\textpm{}\,3.1 & 14.3\,\textpm{}\,12.2 & 2.1\,\textpm{}\,2.4 & 45.7\,\textpm{}\,17.4 & 3.6\,\textpm{}\,3.0 & 45.7\,\textpm{}\,17.4 & 2.9\,\textpm{}\,2.7 \\
     & Nationality & 1.61\,\textpm{}\,0.44 & 48.6\,\textpm{}\,17.4 & 2.9\,\textpm{}\,2.7 & 42.9\,\textpm{}\,17.2 & 3.6\,\textpm{}\,3.0 & 45.7\,\textpm{}\,17.4 & 3.6\,\textpm{}\,3.0 & 25.7\,\textpm{}\,15.2 & 2.9\,\textpm{}\,2.8 \\
     & Birthplace & 4.30\,\textpm{}\,2.05 & 17.1\,\textpm{}\,13.1 & 1.4\,\textpm{}\,2.0 & \textbf{5.7\,\textpm{}\,8.1} & 2.1\,\textpm{}\,2.4 & 25.7\,\textpm{}\,15.2 & 1.4\,\textpm{}\,2.0 & 17.1\,\textpm{}\,13.1 & 1.4\,\textpm{}\,2.0 \\
     & UUID & 5.00\,\textpm{}\,0.62 & \textbf{11.4\,\textpm{}\,11.1} & 1.4\,\textpm{}\,2.0 & 14.3\,\textpm{}\,12.2 & 1.4\,\textpm{}\,2.0 & \textbf{14.3\,\textpm{}\,12.2} & 0.0\,\textpm{}\,0.0 & \textbf{14.3\,\textpm{}\,12.2} & 0.7\,\textpm{}\,1.4 \\
    \midrule
    \multirow{4}{*}{\rotatebox[origin=c]{90}{\shortstack{\textbf{Hubble}\\\textbf{8B-500B}}}} & Email & 0.43\,\textpm{}\,0.42 & 48.6\,\textpm{}\,17.4 & 3.6\,\textpm{}\,3.0 & 40.0\,\textpm{}\,17.1 & 7.1\,\textpm{}\,3.9 & 25.7\,\textpm{}\,15.2 & 4.3\,\textpm{}\,3.2 & 31.4\,\textpm{}\,16.2 & 6.4\,\textpm{}\,3.9 \\
     & Nationality & 1.09\,\textpm{}\,0.31 & 45.7\,\textpm{}\,17.4 & 10.0\,\textpm{}\,4.6 & 25.7\,\textpm{}\,15.2 & 8.6\,\textpm{}\,4.3 & 45.7\,\textpm{}\,17.4 & 8.6\,\textpm{}\,4.3 & 40.0\,\textpm{}\,17.1 & 6.4\,\textpm{}\,3.9 \\
     & Birthplace & 2.30\,\textpm{}\,1.31 & 17.1\,\textpm{}\,13.1 & 3.6\,\textpm{}\,2.9 & 2.9\,\textpm{}\,5.8 & 4.3\,\textpm{}\,3.2 & 28.6\,\textpm{}\,15.7 & 2.9\,\textpm{}\,2.7 & 14.3\,\textpm{}\,12.2 & 3.6\,\textpm{}\,2.9 \\
     & UUID & 5.70\,\textpm{}\,0.70 & \textbf{0.0\,\textpm{}\,0.0} & 4.3\,\textpm{}\,3.3 & \textbf{0.0\,\textpm{}\,0.0} & 5.0\,\textpm{}\,3.6 & \textbf{0.0\,\textpm{}\,0.0} & 4.3\,\textpm{}\,3.3 & \textbf{0.0\,\textpm{}\,0.0} & 5.0\,\textpm{}\,3.4 \\
    \midrule
    \multirow{3}{*}{\rotatebox[origin=c]{90}{\shortstack{\textbf{Hubble}\\\textbf{1B-100B}}}} & Email & 0.71\,\textpm{}\,0.61 & 74.3\,\textpm{}\,15.2 & 8.6\,\textpm{}\,6.4 & \textbf{5.7\,\textpm{}\,8.1} & 2.9\,\textpm{}\,4.0 & 77.1\,\textpm{}\,14.6 & 10.0\,\textpm{}\,6.8 & 80.0\,\textpm{}\,13.9 & 10.0\,\textpm{}\,6.8 \\
     & Nationality & 1.46\,\textpm{}\,0.37 & 62.9\,\textpm{}\,16.8 & 1.4\,\textpm{}\,2.8 & 68.6\,\textpm{}\,16.2 & 0.0\,\textpm{}\,0.0 & 51.4\,\textpm{}\,17.4 & 1.4\,\textpm{}\,2.8 & \textbf{45.7\,\textpm{}\,17.4} & 4.3\,\textpm{}\,4.9 \\
     & UUID & 5.89\,\textpm{}\,0.80 & \textbf{48.6\,\textpm{}\,17.4} & 1.4\,\textpm{}\,2.8 & \textbf{5.7\,\textpm{}\,8.1} & 2.9\,\textpm{}\,4.0 & \textbf{48.6\,\textpm{}\,17.4} & 1.4\,\textpm{}\,2.8 & 51.4\,\textpm{}\,17.4 & 1.4\,\textpm{}\,2.8 \\
    \midrule
    \multirow{3}{*}{\rotatebox[origin=c]{90}{\shortstack{\textbf{Hubble}\\\textbf{1B-500B}}}} & Email & 0.36\,\textpm{}\,0.28 & 51.4\,\textpm{}\,17.4 & 11.4\,\textpm{}\,7.4 & \textbf{8.6\,\textpm{}\,9.8} & 20.0\,\textpm{}\,8.3 & \textbf{48.6\,\textpm{}\,17.4} & 11.4\,\textpm{}\,7.4 & 54.3\,\textpm{}\,17.4 & 7.1\,\textpm{}\,6.0 \\
     & Nationality & 1.09\,\textpm{}\,0.24 & 71.4\,\textpm{}\,15.7 & 2.9\,\textpm{}\,4.0 & 71.4\,\textpm{}\,15.7 & 24.3\,\textpm{}\,9.4 & 71.4\,\textpm{}\,15.7 & 1.4\,\textpm{}\,2.8 & 80.0\,\textpm{}\,13.9 & 1.4\,\textpm{}\,2.8 \\
     & UUID & 4.43\,\textpm{}\,0.61 & \textbf{45.7\,\textpm{}\,17.4} & 0.0\,\textpm{}\,0.0 & \textbf{8.6\,\textpm{}\,9.8} & 1.4\,\textpm{}\,2.8 & 65.7\,\textpm{}\,16.5 & 0.0\,\textpm{}\,0.0 & \textbf{51.4\,\textpm{}\,17.4} & 0.0\,\textpm{}\,0.0 \\
    \bottomrule
  \end{tabular*}
\end{table}


\begin{table}[!htbp]
  \centering
  \setlength{\belowcaptionskip}{6pt}
  \caption{\textbf{Mean retain damage across methods.} Accuracy decrease (percentage points), averaged over GA, GD, NPO, and DPO, with 95\% $t$ intervals across methods. $\Delta$ is the difference between in-relation and out-relation damage.}
  \label{tab:yago-method-mean}
  \small
  \renewcommand{\arraystretch}{1.12}
  \begin{tabular}{@{}llrrr@{}}
    \toprule
    Model & Relation & In-relation & Out-relation & $\Delta$ \\
    \midrule
    Hubble-8B-100B & Email & 45.7\,\textpm{}\,40.8 & 3.0\,\textpm{}\,1.1 & 42.7\,\textpm{}\,39.9 \\
     & Nationality & 40.7\,\textpm{}\,16.3 & 3.2\,\textpm{}\,0.7 & 37.5\,\textpm{}\,16.1 \\
     & Birthplace & 16.4\,\textpm{}\,13.1 & 1.6\,\textpm{}\,0.6 & 14.8\,\textpm{}\,13.6 \\
     & UUID & 13.6\,\textpm{}\,2.3 & 0.9\,\textpm{}\,1.1 & 12.7\,\textpm{}\,3.0 \\
    \midrule
    Hubble-8B-500B & Email & 36.4\,\textpm{}\,15.9 & 5.4\,\textpm{}\,2.7 & 31.1\,\textpm{}\,16.6 \\
     & Nationality & 39.3\,\textpm{}\,15.0 & 8.4\,\textpm{}\,2.3 & 30.9\,\textpm{}\,14.8 \\
     & Birthplace & 15.7\,\textpm{}\,16.8 & 3.6\,\textpm{}\,0.9 & 12.1\,\textpm{}\,17.7 \\
     & UUID & 0.0\,\textpm{}\,0.0 & 4.6\,\textpm{}\,0.7 & -4.6\,\textpm{}\,0.7 \\
    \midrule
    Hubble-1B-100B & Email & 59.3\,\textpm{}\,57.0 & 7.9\,\textpm{}\,5.4 & 51.4\,\textpm{}\,51.6 \\
     & Nationality & 57.1\,\textpm{}\,16.6 & 1.8\,\textpm{}\,2.9 & 55.4\,\textpm{}\,19.1 \\
     & UUID & 38.6\,\textpm{}\,34.9 & 1.8\,\textpm{}\,1.1 & 36.8\,\textpm{}\,36.1 \\
    \midrule
    Hubble-1B-500B & Email & 40.7\,\textpm{}\,34.3 & 12.5\,\textpm{}\,8.6 & 28.2\,\textpm{}\,42.6 \\
     & Nationality & 73.6\,\textpm{}\,6.8 & 7.5\,\textpm{}\,17.8 & 66.1\,\textpm{}\,21.3 \\
     & UUID & 42.9\,\textpm{}\,38.8 & 0.4\,\textpm{}\,1.1 & 42.5\,\textpm{}\,39.8 \\
    \bottomrule
  \end{tabular}
\end{table}

\FloatBarrier
\subsubsection{Correlation Analysis of Memorization Floor and Retain Damage}
\label{app:factual-cor}
Table~\ref{tab:yago-correlations} relates each relation's mean memorization floor to its in-relation damage, separately for every model and method.
Fifteen of the sixteen Pearson coefficients are negative, so a higher floor goes with less damage at both model sizes and both pretraining budgets.
The association is stronger in Hubble-8B, where $r$ runs from $-0.978$ to $-0.517$, than in Hubble-1B, where it runs from $-0.948$ to $+0.435$; the single positive value is NPO in Hubble-1B-500B.

\begin{table}[!htbp]
  \centering
  \setlength{\belowcaptionskip}{6pt}
  \caption{\textbf{Memorization floor--damage correlations in factual recall.} Pearson\textquotesingle s $r$ and Spearman\textquotesingle s $\rho_s$ over the relations of one model and method: four for Hubble-8B and three for Hubble-1B. With three relations $\rho_s$ can only take the values $0$, $\pm0.5$ and $\pm1$, so the Hubble-1B rank coefficients are coarse.}
  \label{tab:yago-correlations}
  \small
  \begin{tabular}{@{}llrr@{}}
    \toprule
    Model & Method & $r$ & $\rho_s$ \\
    \midrule
    Hubble-8B-100B & GA & $-$0.966 & $-$1.000 \\
     & GD & $-$0.517 & $-$0.316 \\
     & NPO & $-$0.978 & $-$0.949 \\
     & DPO & $-$0.889 & $-$1.000 \\
    \midrule
    Hubble-8B-500B & GA & $-$0.943 & $-$1.000 \\
     & GD & $-$0.829 & $-$1.000 \\
     & NPO & $-$0.839 & $-$0.400 \\
     & DPO & $-$0.908 & $-$0.800 \\
    \midrule
    Hubble-1B-100B & GA & $-$0.948 & $-$1.000 \\
     & GD & $-$0.379 & \phantom{$-$}0.000 \\
     & NPO & $-$0.681 & $-$1.000 \\
     & DPO & $-$0.481 & $-$0.500 \\
    \midrule
    Hubble-1B-500B & GA & $-$0.538 & $-$0.500 \\
     & GD & $-$0.347 & \phantom{$-$}0.000 \\
     & NPO & \phantom{$-$}0.435 & \phantom{$-$}0.500 \\
     & DPO & $-$0.431 & $-$0.500 \\
    \bottomrule
  \end{tabular}
\end{table}

\FloatBarrier
\section{Directing Suppression}
\label{app:redistribution}

\subsection{BMA Redistribution Experiment}
\label{app:bma-redistribution}
\paragraph{Configuration.}
The comparison runs on two task configurations, with inputs $a,b\in\{0,\ldots,P-1\}$ and $PM$ output classes, each a one-layer transformer trained on 50\% of the input pairs for 40,000 epochs: $(P,M)=(131,2)$, giving 262 output classes, with model and MLP widths 128 and 512 and weight decay 1; and $(277,4)$, giving 1,108 classes, with widths 256 and 1,024 and weight decay 3.
Neither model is one of those swept in Appendix~\ref{app:bma}, whose output space is fixed at 256.
We unlearn 10\% of the training pairs with full-batch AdamW at learning rate $10^{-4}$, zero weight decay, and no retain term, evaluating every ten updates, and the training run, split, and label assignment are fixed within a configuration.

\paragraph{Conditions.}
Both conditions are RL and differ only in where the removed probability is allowed to go.
The unrestricted target is drawn uniformly from every answer other than $y^\star$, while the within-bucket target is drawn uniformly from the other $M-1$ answers of $y^\star$'s own bucket and therefore still satisfies the modular rule.
Targets are drawn once and held fixed, so the intervention changes the set of permitted replacements and nothing else.

\paragraph{Results.}
Conditions are compared at 10\% forget exact accuracy, interpolated linearly between the first pair of recorded evaluations that brackets it; this accuracy criterion differs from the mean correct-answer probability used for the sweep in Section~\ref{sec:excess-prediction}.
At that point within-bucket targets preserve far more retain exact accuracy---75.7\% against 15.4\% at $(P,M)=(131,2)$ and 71.3\% against 28.0\% at $(277,4)$---and they leave forget bucket accuracy near 100\%, where unrestricted targets drop it to 12.5\% and 14.7\% (Table~\ref{tab:bma-redistribution}).


\begin{table}[htbp]
  \centering
  \setlength{\belowcaptionskip}{6pt}
  \caption{\textbf{Within-bucket redistribution preserves retain performance.} RL with unrestricted or within-bucket incorrect targets at 10\% forget exact accuracy. Entries report retain exact and forget bucket accuracies (\%) from one trained model and split per setting.}
  \label{tab:bma-redistribution}
  \small
  \begin{tabular}{@{}rrlrr@{}}
    \toprule
    $P$ & $M$ & Relabeling target & Retain & Forget bucket \\
    \midrule
    131 & 2 & Unrestricted & 15.4 & 12.5 \\
    131 & 2 & Within bucket & 75.7 & 99.9 \\
    \midrule
    277 & 4 & Unrestricted & 28.0 & 14.7 \\
    277 & 4 & Within bucket & 71.3 & 99.7 \\
    \bottomrule
  \end{tabular}
\end{table}

\subsection{Implications for Existing Unlearning Objectives}
\label{app:redistribution-llm}
LLMs have no explicit modular bucket, but the linguistic and semantic regularities studied in Section~\ref{sec:llm} suggest possible analogues.
Within a forget document the two are interleaved: some tokens follow grammatical and contextual constraints shared with other text, while others carry the sample-specific wording that only memorization supplies.
The redistribution experiment preserves the shared rule by controlling where the removed probability goes; the same principle can be pursued by controlling which predictions are suppressed in the first place.
Token-wise loss regularizers already set how strongly each token is suppressed, and a reference model without the targeted examples can indicate which tokens the shared mechanisms already predict.
We therefore ask whether regularizers that put more weight on the tokens the reference model does not already predict preserve more of the retain set.
They do: across the eight token weightings we compare, the sharper this separation is at the point where unlearning stops, the more retain performance survives.
Because each objective is reduced to its token weight, the comparison speaks to that weighting rather than to the published methods as a whole.

\paragraph{Setup.}
We unlearn the perturbed Hubble-1B-100B checkpoint.
Of the 80 Gutenberg Popular passages inserted 256 times into its pretraining data, a random half forms the forget set and the other half the retain set, and results are averaged over three random splits.
Each regularizer contributes only its token weight $\omega_t$ (Table~\ref{tab:tokw-defs}) applied to a common gradient-ascent loss; the other components of the original objectives, such as retain terms, are not used.
All regularizers share the same optimizer and stop when $\bar{P}_f$ first reaches 0.1, the criterion of Appendix~\ref{app:verbatim-config}.

\paragraph{Measuring separation.}
We estimate the memorization floor of each forget token as in Section~\ref{sec:verbatim}, taking the perturbed model being unlearned as the target and the standard model as the reference, and split the tokens at one bit: those above the threshold are still predicted largely through memorization, while those below it are already carried by the constraints shared with other text.
The threshold is ours rather than one taken from prior work: one bit means that the perturbed model assigns the token at least twice the probability the standard model does, which we take as a simple cutoff for a substantial memorization component.
Separation is the AUC of a regularizer's weights between the two groups, that is, the probability that a memorization-driven token receives a larger weight than one the shared mechanisms already predict, with ties counted as one half; uniform weighting, as in GA, gives $0.5$.

\paragraph{Result.}
Retain damage arises when suppression keeps acting on tokens the model no longer predicts through memorization, and such tokens accumulate as unlearning proceeds: they grow from about a quarter of the forget tokens at the start to about half at the stopping point.
We therefore read the AUC at the stopping point, where it reflects whether a regularizer spares them, and compare it with retain performance at the same point.
Regularizers that separate the two groups more sharply preserve the retain set better (Spearman $\rho=0.76$ across the eight regularizers; Table~\ref{tab:tokw-auc}), whereas the AUC measured at the start of unlearning shows no such trend ($\rho=-0.30$).

\begin{table}[htbp]
\centering
\caption{\textbf{Token-wise regularizers compared in Appendix~\ref{app:redistribution-llm}.} $p_t$: the current model's probability of the gold token; $p_t^{\mathrm{ref}}$: the same before unlearning; $H_t$: predictive entropy at temperature $T=1.5$; $n_x$: document length. Exponents are hyperparameters: $\beta=5$ for WGA, $4$ for TNPO and $1$ for Imp; $(\beta_1,\beta_2)=(5,1)$ for SatImp.}
\label{tab:tokw-defs}
\vspace{0.5em}
\small
\renewcommand{\arraystretch}{1.15}
\begin{tabular}{@{}lll@{}}
\toprule
Method & Weight $\omega_t$ & Tokens weighted most \\
\midrule
GA & $1$ & all equally \\
WGA \citep{wang2025rethinking} & $p_t^{\beta}$ & tokens the model still predicts confidently \\
TNPO \citep{wang2025rethinking} & $2p_t^{\beta}/\bigl(p_t^{\beta}+(p_t^{\mathrm{ref}})^{\beta}\bigr)$ & tokens whose probability has dropped least \\
Imp \citep{yang2025exploring} & $(1-p_t)^{\beta}$ & tokens the model predicts poorly \\
SatImp \citep{yang2025exploring} & $p_t^{\beta_1}(1-p_t)^{\beta_2}$ & confident but not yet saturated tokens \\
ETW \citep{koh2026forgetmattersrestselective} & $n_x H_t/\sum_s H_s$ & tokens with uncertain predictions \\
SCN \citep{dong-etal-2025-undial} & $\mathbf{1}[\text{noun}]$ & nouns and proper nouns (spaCy) \\
SCE \citep{dong-etal-2025-undial} & $\mathbf{1}[\text{named entity}]$ & named entities (spaCy) \\
\bottomrule
\end{tabular}
\end{table}


\begin{table}[!t]
\centering
\caption{\textbf{Regularizers that better separate tokens above from those below the one-bit memorization floor (higher AUC) retain more.} Both measured when the forget probability reaches 0.1; mean $\pm$ 95\% CI over three seeds.}
\label{tab:tokw-auc}
\vspace{0.5em}
\small
\begin{tabular}{lcc}
\toprule
Method & AUC & Retain \\
\midrule
SatImp & 0.65\,\textpm{}\,0.05 & 0.236\,\textpm{}\,0.007 \\
WGA & 0.61\,\textpm{}\,0.07 & 0.246\,\textpm{}\,0.019 \\
TNPO & 0.61\,\textpm{}\,0.06 & 0.243\,\textpm{}\,0.014 \\
ETW & 0.55\,\textpm{}\,0.04 & 0.199\,\textpm{}\,0.013 \\
GA & 0.50\,\textpm{}\,0.00 & 0.217\,\textpm{}\,0.015 \\
SCE & 0.45\,\textpm{}\,0.00 & 0.191\,\textpm{}\,0.006 \\
SCN & 0.42\,\textpm{}\,0.02 & 0.223\,\textpm{}\,0.007 \\
Imp & 0.39\,\textpm{}\,0.03 & 0.187\,\textpm{}\,0.016 \\
\midrule
Spearman $\rho$ & \multicolumn{2}{c}{0.76} \\
\bottomrule
\end{tabular}
\end{table}

\end{document}